\documentclass[a4paper,10pt]{article}
\usepackage{a4wide}
\usepackage{setspace}
\usepackage{xspace}
\usepackage{algorithm,algpseudocode}
\usepackage{url}
\usepackage{graphicx}
\providecommand{\keywords}[1]{\textbf{\textit{Keywords}} #1}

 \usepackage[pagewise]{lineno}

\usepackage{algorithm,algpseudocode}
\usepackage{tikz}
\usepackage{array}
\usepackage{tabularx}
\usepackage[title]{appendix}
\usepackage[T1]{fontenc}
\usepackage{multirow}
\usepackage[utf8]{luainputenc}
\usepackage{graphicx}
\usepackage{subfig}
\usepackage{captdef}
\usepackage{float}
\usepackage{hyperref}
\usepackage{amsmath, amsthm, amssymb, bbm}
\providecommand{\keywords}[1]{\textbf{Keywords:  } #1}
\providecommand{\classifications}[1]{\textbf{
Mathematics Subject Classification:} #1}

\newtheorem{thm}{Theorem}[section]
\newtheorem{prop}{Proposition}[section]
\newtheorem{corollary}{Corollary}[section]

\newtheorem{lem}{Lemma} [section]

\newcommand{\R}{\mathbb{R}}
\newcommand{\T}{\mathbb{T}}

\newcommand{\Rn}{\mathbb{R}^n}
\newcommand{\MATLAB}{\textsc{Matlab}\xspace}

\title{A diffusion-map-based algorithm for gradient computation on manifolds and applications}

\author{Alvaro Almeida Gomez $^1$ \\ email \href{mailto:alvaro.gomez@ku.ac.aem}{alvaro.gomez@ku.ac.ae} 
   \and Antônio J.  Silva Neto $^2$ \\ email \href{mailto:ajsneto@iprj.uerj.br}{ajsneto@iprj.uerj.br} 
   \and Jorge P. Zubelli $^1$\\ email \href{mailto:zubelli@gmail.com}{zubelli@gmail.com}}

   \date{%
    $^1$Khalifa University, P.O. Box 127788, Abu Dhabi, United Arab Emirates\\%
    $^2$IPRJ-UERJ, R. Bonfim 25, Nova Friburgo 28625-570, Brazil\\[2ex]%
    \today
}

\makeatletter
\numberwithin{equation}{section}
\numberwithin{figure}{section}

\begin{document}

\maketitle
\begin{abstract}
We present a technique to estimate the Riemannian gradient of a given function defined on interior points of a Riemannian submanifold in the Euclidean space based on a sample of function evaluations at points in the submanifold. This approach is based on the estimates of the Laplace-Beltrami operator proposed in the diffusion-map theory. Analytical convergence results of the  Riemannian gradient expansion are proved.
\textcolor{black}{The methodology provides a new algorithm to compute the gradient in cases where classical methods for numerical derivatives fail. For instance, in classification problems, and in cases where the information is provided in an unknown nonlinear lower-dimensional submanifold lying in high-dimensional spaces.
The results obtained in this article connect the theory of diffusion maps with the theory of learning gradients on manifolds.
We apply the Riemannian gradient estimate in a gradient-based algorithm providing a derivative-free optimization method.}
We test and validate several applications, including tomographic reconstruction from an unknown random angle distribution, and the sphere packing problem in dimensions 2 and 3.

\end{abstract}
\\
\classifications {Primary: 49N45, 65K05, 90C53, 65J22; Secondary: 94A08, 68T01, 68T20.}

\section{Introduction}

A vast number of iterative minimization algorithms rely on the fact that the negative gradient determines the steepest descent direction. The applications in science, in general, and inverse problems, in particular, abound~\cite{ denoising,gradteo2,paperfundamental}. Some examples of these algorithms are the Gradient Descent and Newton's method~\cite{claudia} which have deep theoretical aspects~\cite{smale, shub, yuan,benarsvaiter}. Although most of the focus in applications concern Euclidean spaces, these methods are also important in the context of  Riemannian geometry. See~\cite{shub2,sepul,edelman,smith,bloch} and references therein. 

In this article, we address an important task in the aforementioned methods, namely to compute the Riemannian gradient from data or from inexactly computed function values. In many cases such gradient is not easily computable due to the complexity of the function's local behavior. Problems also arise whenever the available information consists of high-dimensional unsorted sample points lying in an unknown nonlinear lower-dimensional submanifold~\cite{muksa}. The latter issue does not allow the tangent space to be efficiently and economically computed from noisy sample points. 
Thus, one of the purposes of this article is to confront such difficulties. We emphasize that we focus on giving Riemannian gradient estimates instead of proposing an optimization method. In other words, we compute approximations of the Riemannian gradient of a function using sample points. An important feature of our approximations is that it does not depend on differential conditions of the function.  The main tool to compute these estimates is the diffusion-map theory. The latter is a dimensionality reduction methodology that is based on the diffusion process in a manifold. See Refs.~\cite{COIFMAN20065,COIFMAN201479,Coifman7426} for more details.   

An important feature of the theory of diffusion maps is that it recovers the  Laplace-Beltrami operator when the dataset approximates a Riemannian submanifold of $\Rn$. The diffusion-map theory is based on a symmetric kernel defined on the dataset. The symmetric kernel measures the connectivity between two points. Our approach is based on implementing this theory in the recently developed case of asymmetric kernels~\cite{alvaro}. Compared to symmetric kernels, asymmetric kernels provide more details on how the information is distributed in each direction. This characteristic allows us to know the path with the greatest variations.

\textcolor{black}{In comparison with classical methods where the gradient is numerically computed using the knowledge of the differential structure of the manifold, our approach focuses on cases where the available information consists only of sample points lying in an unknown manifold. In a certain sense, we follow the 
paradigm of a data driven computation to solve the problem in the spirit of \cite{GLZ2018}.
}

\textcolor{black}{
The problem we consider here appears, for instance, in the context  
of  the Learning Gradient Theory~\cite{learning2}. In this framework, one computes the gradient of a function defined on a submanifold and apply it to  supervised learning, in algorithms for classification, and dimensionality reduction .}

\textcolor{black}{ However, the estimates in the Learning Gradient Theory are based on the representation theorem for Reproducing Kernel Hilbert Space (RKHS), which requires solving an optimization problem to compute the coefficients in the representation. This, in turn,  might be computationally expensive when the sample size is large enough. In the present work, we use the diffusion-map theory  and the family of associated kernels to give a closed form for the gradient approximation, thus, improving the computational complexity.}
 As an application of our methodology, we use our approach as the main direction in a gradient-based algorithm. See Ref.~\cite{sepul}. The main advantage of using this operator is that it does not depend on some {\it a  priori} knowledge of the Riemannian gradient of the function. Furthermore, since the operator is defined as an integral, then it is robust to noise in the data.

We test our proposed gradient-based algorithm in two applications. 
Firstly, we apply it to the sphere packing problem in dimensions $2$ and $3$. This problem was addressed numerically, 
in Ref.~\cite[Chapter 2]{spheregrassmanian}. Here, an optimization algorithm using the  gradient descent technique is proposed to tackle the sphere packing problem on a Grassmannian manifold, in this case, there is a closed form to compute the gradient of the function.
\textcolor{black}{In contradistinction, in the present article, as an experiment, we consider the sphere packing in the Euclidean space. This is more difficult because there is no closed form for the gradient of the objective function
due to the singularities in the ambient space. In fact, the objective function is not differentiable. In our approach, we reformulate the sphere packing problem as an optimization problem over the special linear group, and we use the proposed  methodology  to find a computational solution.}
\textcolor{black}{To analyze the performance of the methodology, we test and compare the proposed algorithm with  the derivative-free solvers (\textsc{PSO} and \textsc{Nelder-Mead}) implemented in the \textsc{Manopt toolbox}, described in Refs. \cite{manopt1,manopt2}.}

Secondly, we apply the proposed methodology to the tomographic reconstruction problem from samples of unknown angles.
This post-processing algorithm is parallelizable. It also has a similar flavor to the algorithm developed in  Refs. \cite{MZ2001,MR2002m} since we are trying to solve a high dimensional optimization problem with a swarm of computed auxiliary data. In the latter case, this is done with the approximation to the roots of a high-degree polynomial. 
Our reconstruction method is based on using the diffusion maps for a partition of the dataset, instead of considering the complete database as proposed in Ref. \cite{angucoif}. We remark that we reconstruct the image except for a possible rotation and reflection.
Compared to traditional reconstruction methods Refs. \cite{angucoif,angudesco}, our method does not assume the hypothesis that the distribution of the angles is previously known, which makes it a more general and practical method for numerical implementations. In addition,  our method runs faster and more efficiently than the method proposed in Ref. \cite{angucoif}. In fact, if the number of sample points is  $us+r$ with $ r < s < u$, then the complexity of the algorithm proposed  in Ref. \cite{angucoif}  is $O(u^3 \, s^3)$, while our algorithm runs with complexity $O(u\, s^3)$. On the other hand, the numerical implementation described in Ref. \cite{india} of the methodology proposed in Ref. \cite{angudesco}, uses brute force which is not suitable when the number of sample points is large.
\par  This paper is organized as follows, in Section~\ref{diffusion}, we give a brief exposition of the classical representation theory for diffusion distances proposed in Refs.~\cite{COIFMAN20065, COIFMAN201479, Coifman7426}, and we state our main result in Theorem~\ref{teo1}.
In Section~\ref{gradientflowsect}, we review facts about flows defined over manifolds, and we show how to use the flow generated by the approximations to find minimizers. In Section~\ref{applications}, we show some experiments related to the sphere packing problem, and we also show the effectiveness of our tomographic reconstruction method when the angles are unknown. Finally, in Appendices~\ref{ape1} and \ref{ape2}, we cover the technical details of the proof of the main result.

\section{Diffusion-Maps}
\label{diffusion}
In this section, we review some facts on diffusion-map theory. We refer the reader to Refs.~\cite{COIFMAN20065,COIFMAN201479,Coifman7426} for more details. Diffusion-maps is a nonlinear dimensionality reduction method that is based on the diffusion process over datasets.
In diffusion-map theory, we assume that our dataset $X=\{{x_i} \}_{i=1}^{k}$ satisfies $X \subset \mathcal{M}  \subset\Rn$, where $\mathcal{M}$ is a $d-dimensional$ Riemannian submanifold of the ambient space $\Rn$.  In this case the dimension $d$ of   $\mathcal{M}$  is assumed to be much smaller than $n$.
In our approach, we use  asymmetric vector-valued kernels as in Ref.~\cite{alvaro}. The main advantage of using these kernels is that we have a more specific description of the distribution of the dataset in certain directions. Based on the expansion for the Laplace-Beltrami operator proposed in Ref.~\cite{COIFMAN20065} we  recover the Riemannian gradient. Firstly, we consider  the vector-valued kernel
$$ \overline{K}_t:\mathcal{M} \times \mathcal{M} \to \R^n ,$$
defined as
$$  \overline{K}_t(x,y)= (y-x) e^{\frac{-\| y-x \|^2}{2 t^2}}. $$
We fix the exponent $\delta \in (1/2,1)$,
and let $d_t(x)$ be defined by
$$d_t(x)=\int_{U(x,t^\delta)} e^{\frac{-\| y-x \|^2}{2 t^2}} dy,$$
where 
\begin{equation}
    \label{conjuntopequ}
    U(x,t)=\{ y\in \mathcal{M} | \|y-x\| \le t \}.
\end{equation}
Here, the parameter $\delta$ has to be in $(1/2,1)$ to guarantee convergence of the estimates as shown in Lemma \ref{lemaprinci}. We consider  the Markov normalized kernel given by
$$ \rho_t (x,y) = \frac{\overline{K}_t(x,y)}{d_t(x)}   .$$
For a function $f$, we define the operator 
\begin{equation}
    \overline{P}_{t}f (x)=\int_{U(x,t^\delta)}  \rho_t(x,y) (f(y)-f(x)) dy.
    \label{kerneloperator}
\end{equation}
We now show that this operator approximates the Riemannian gradient of a given function on some Riemannian submanifold.  The technical details of the proof are given in Appendices~\ref{ape1} and~\ref{ape2}.
\begin{thm} \label{teo1}
Let $\mathcal{M}$  be a   Riemannian submanifold of $\Rn$ and assume that the function $f$ is smooth, and  $x$ is an interior point of $\mathcal{M}$. Then, the following estimate holds

\begin{equation}
  \overline{P}_{t} f (x) =\nabla f(x) \, t^{2}+O(t^{4 \delta }), 
\end{equation}

where $\nabla f$ is the Riemannian gradient of $f$. In particular, we have that
\begin{equation}
    \lim_{t\to 0}\frac{\overline{P}_{t} f (x)}{ t^{2}} =\nabla f(x). 
    \label{ecuacionprincipal}
\end{equation}

\end{thm}
Note that the operator $\overline{P}_{t} $ does not depend on differentiability conditions. Furthermore, since the operator is defined as an integral one, then it is robust to noise perturbation. Considering these characteristics, we use this operator as a substitute for the Riemannian gradient as the main direction of a gradient-based algorithm on manifolds detailed in Ref.~\cite{sepul,sato}.

\section{Flows and optimization methods on submanifolds}
\label{gradientflowsect}

In this section, we review some facts about flows defined on submanifolds and we show how the flow generated by the vector field $\overline{P}_{t} f(\cdot)$   can be used in optimization methods.
\par
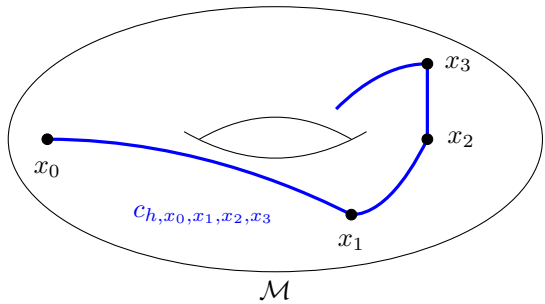
\begin{figure}[htp] 
    \centering
\begin{tikzpicture}
\draw (-1,0) to[bend left] (1,0) ;
  \draw (-1.2,.1) to[bend right] (1.2,.1) ;
  \draw[rotate=0] (0,0) ellipse (100pt and 50pt) node [below=50pt]{$\mathcal{M} $};
\draw [very thick,blue](-3,0)  parabola (1,-1) node [left=25pt] {$c_{h,x_0,x_1,x_2,x_3}$} ;
\draw [very thick,blue] (1,-1)  parabola (2,0);
\draw [very thick,blue](2,0)  parabola (2,1);
\draw [very thick,blue] (2,1)  parabola (0.8 ,0.4);
\filldraw[black] (-3,0) circle (2pt) node[below=5pt] {$x_0$};
\filldraw[black] (1,-1) circle (2pt) node[below=4pt] {$x_1$};
\filldraw[black] (2,0) circle (2pt) node[right=4pt] {$x_2$};
\filldraw[black] (2,1) circle (2pt) node[right=3pt] {$x_3$};
\end{tikzpicture}
\caption{Piecewise curve  obtained by concatenating four curves.}
\label{concadenaop}
\end{figure}

 Assume that $h:\mathcal{M} \to \R^n$ is a continuous function defined on the submanifold $\mathcal{M} \subset \Rn$. We say that a curve $b$ starts at $x_0$, if  $b(0) = x_0$. The Peano existence theorem guarantees that for all $x_0 \in \mathcal{M}$, there exists a smooth curve $c_{h,x_0}:(-\varepsilon, \varepsilon) \to \mathcal{M}$ starting at $x_0$, which is solution of
\begin{equation}
  \label{gradientflowequ}
  \begin{aligned}
    c_{h,x_0}'(s) &= - h(c_{h,x_0}(s)).\\        
  \end{aligned}
\end{equation}
 We refer the reader to Ref.~\cite{libroedo} for a complete background about ordinary differential equations. We observe that assuming only the continuity condition, the uniqueness of the curve is not guaranteed. Since the solution of Eq.~(\ref{gradientflowequ}) may not be unique, we can concatenate solutions as follows. Let $ c_{h,x_0}$ be a solution of Eq.~(\ref{gradientflowequ}) starting at the point $x_0$. For a fix $s_1$ in the domain of $ c_{h,x_0}$, we define $x_1=c_{h,x_0}(s_1)$. If $ c_{h,x_1}$ is a solution of Eq.~(\ref{gradientflowequ}) starting in $x_1$,  we define a new curve $c_{h,x_0,x_1}$ as
$$ c_{h,x_0,x_1}(s) = \left\{\begin{array}{lr}
        c_{h,x_0}(s), & \text{for }  s \le s_1\\
        c_{h,x_1}(s-s_1), & \text{for } s_1< s\\ 
\end{array} \right. \mbox{. }$$
Proceeding recursively, we obtain a piecewise differentiable curve $c_{h,x_0,x_1,x_2 \cdots}(s)$ starting at $x_0$, and satisfying Eq.~(\ref{gradientflowequ}) (except in a discrete set). See Figure \ref{concadenaop} for a graphic description. In this case, we say that the curve $c_{h,x_0,x_1,x_2 \cdots}(s)$  is a piecewise solution of Eq.~(\ref{gradientflowequ}). We focus on curves which are solutions (except in a discrete set) of Eq.~(\ref{gradientflowequ}), because these curves allow updating the direction in which we look for stationary points.

Suppose that $f:\mathcal{M} \to \R$ defines a smooth function. In this case we consider the vector field $h=\nabla f$. If $c_{h,x_0,x_1,x_2 \cdots}$ is a piecewise solution of  Eq.~(\ref{gradientflowequ}) starting at $x_0$, then,  for all $t$ (except in a discrete set), we have that
\begin{equation}
     \| c_{h,x_0}'(s) \|^2 = - \frac{d}{ds} f (c_{h,x_0}(s)).
     \label{energiagradiente}
\end{equation}
Therefore, the function $ f (c_{h,x_0}(\cdot))$ is decreasing. \textcolor{black}{Thus, we can use use the flow $c$ to find a local minimum for the function $f$ .}

\subsection{Lipschitz  functions}
We recall that $f$ is a locally Lipschitz function if for all $x \in \mathcal{M}$  there exists a neighborhood $x \in U\subset \mathcal{M}$ and a positive constant $C$, such that for  all  $y \in U$ it holds that
$$ | f(x)-f(y)| \le C \|x-y\|_{L^2}.$$
We also recall that the Sobolev space $H^{1}(0,T,\mathcal{M})$ is defined as the set of all square integrable functions from $[0,T]$ to $\mathcal{M}$ whose weak derivative has also finite $L^2$ norm.
 \par Our goal is to use the gradient approximation in Theorem \ref{teo1}   to find minimal points of locally Lipschitz functions. Recall that Rademacher's theorem states that for a locally Lipschitz function $f$, the gradient operator $ \nabla f$ exists almost everywhere. See Ref.~\cite{evans10} for more details. However, for a locally Lipschitz function $ f $, the gradient $\nabla f$ may not exist for all points. In this case, it is not possible to define the gradient flow.
\par To address this  problem, we propose to use the flow generated with $ {\overline{P}_{t}f (x)}$ defined in Eq.~(\ref{kerneloperator}) instead of the gradient. The operator $\overline{P}_{t}f$ is defined as an integral, and thus it is continuous. This fact guarantees the existence of a flow associated with $ \frac{\overline{P}_{t}f (x)}{t^2}$ for arbitrarily small positive $t$.
\par  Now we show that at the points where the function is smooth, this flow approximates a curve for which the function decreases with time. To do that, we first prove a technical result.
\begin{prop}
Suppose that $f$ is continuously differentiable in an open neighborhood of $x_0$. We define the function $J:[0,T] \times \overline{B(x_0, R) \cap \mathcal{M}} \to \Rn$ as
$$ J(t,x) = \left\{\begin{array}{lr}
       \frac { {P}_{t}f (x)} {t^2}, & \text{for }  t > 0\\
        \nabla f (x), & \text{for } t=0\\
  \end{array} \right. \mbox{, }$$
where $B(x_0, R)$ is the ball in $\Rn$ with center $x_0$ and radius $R$. Then, for small enough numbers $T$ and $R$, the function $J$ is uniformly continuous. In particular, there exists a positive constant $M$ such that for all $(t,x) \in (0,T] \times \overline{B(x_0, R) \cap \mathcal{M}}$ the following estimate holds.
\begin{equation}
    \label{estimaflow}
    \frac{ {P}_{t}f (x)} {t^2} \le M.
\end{equation}
\begin{proof} Since the set $ [0,T] \times \overline{B(x_0, R) \cap \mathcal{M}}$ is compact, it is enough to show that $J$ is continuous. Firstly, we show that $J$ is continuous on $(0,T] \times \overline{B(x_0, R) \cap \mathcal{M}}$. For that, we claim that for a continuous vector-valued function $\omega: (0,T] \times \overline{B(x_0, R) \cap \mathcal{M}}  \times \overline{B(x_0, R) \cap \mathcal{M}} \to \R^m $, the operator
$$ \Omega(t,x)= \int_{U(x,t^\delta)} \omega(t,x,y) \, dy,$$
is continuous. In fact, we observe that 
\begin{equation}
    \label{omegagra}
    \Omega(t,x)-\Omega(t_1,x_1)= \int_{U(x,t^\delta)}\omega(t,x,y)-\omega(t_1,x_2,y) \,  dy + \int_{G(t,t_1,x,x_1)}  \omega(t_1,x_2,y) dy,
\end{equation}
where 
$$G(t,t_1,x,x_1)= U(x_1,t_{1}^\delta) \setminus U(x,t^\delta) \cup U(x,t^\delta) \setminus U(x_1,t_{1}^\delta).$$
On the other hand, a straightforward computation shows that 
$$\lim_{(t_1,x_1) \to (t,x)} 1_ {G(t,t_1,x,x_1)} =0,$$
where the convergence is pointwise almost everywhere, therefore
\begin{equation}
    \label{funclimi1}
    \lim_{(t_1,x_1) \to (t,x)} \int_{G(t,t_1,x,x_1)} \omega(t_1,x_2,y) dy=0.
\end{equation}
In addition, since  the function $\omega$ is continuous, then 
\begin{equation}
    \label{funclimi2}
    \lim_{(t_1,x_1) \to (t,x)} \int_{U(x,t^\delta)}\omega(t,x,y)-\omega(t_1,x_2,y) \,  dy=0.
\end{equation}
Using  Eqs.~\eqref{funclimi1} and \eqref{funclimi2} in Eq.~\eqref{omegagra}, we conclude that $\Omega$ is a continuous function. We apply the previous result to the function $w_1(t,x,y)=e^{\frac{-\| y-x \|^2}{2 t^2}}$ to obtain that  $\Omega_1(t,x)=d_t(x)$ is a continuous function. This implies that the function
$$ w_2(t,x,y)= \frac {\rho_t(x,y) (f(y)-f(x))} { t^2},$$
is continuous  on $(0,T] \times \overline{B(x_0, R) \cap \mathcal{M}}$.
Again, we apply the same result to the function $$ w_2(t,x,y),$$ to conclude that $J(t,x)$ is a continuous function on $(0,T] \times \overline{B(x_0, R) \cap \mathcal{M}}$.
\par Moreover, using  Estimate~(\ref{ecuafinal}) of the proof of Theorem~\ref{ecuacionprincipal} and Lemma~\ref{lemma2}, we conclude that the function $J$ is continuous for all points of the form $(0,x)$. This proves our result.
\end{proof}
\label{propauxi}
\end{prop}
\par The estimate of Proposition~\ref{propauxi} states that for a fixed $x_0$, and small $T$, the family of curves   $\{c_{h(t_n),x_0}\}_{t_n}$ is uniformly bounded on the Sobolev space $H^{1}(0,T,\mathcal{M})$. Thus, the Rellich-Kondrachov theorem states that for any sequence $t_n \to 0$, there exists a subsequence $t_{n_k} \to  0$  such that  $c_{h(t_{n_k}),x_0}$ converges  to some curve $c$ in the $L^2$-norm. Observe that by the Arzela-Ascoli theorem, we can also suppose that the sequence  $c_{h(t_n),x_0}$ converges uniformly to $c$. Finally we prove the  main result in this section.
\begin{prop} Assume the same assumptions and notations of Proposition~\ref{propauxi}. Then,  for $t_1<t_2$ we have that
 $$ f(c(t_1))\ge f(c(t_2)). $$
\begin{proof}
 We claim that  $\frac{\overline{P}_{t}f (c_{h(t_n),x_0}(\cdot))}{t^2}$ converges pointwise to $\nabla f (c(\cdot))$, where $ c $ is the curve previously described. In fact, for all $s,$ we have  by Proposition~\ref{propauxi} that
 $$ \lim_{n \to \infty}\frac{\overline{P}_{t}f (c_{h(t_n),x_0}(s))}{t^2}-\nabla f (c_{h(t_n),x_0}(s))=0.$$
 The continuity of the gradient guarantees that
 $$\lim_{n \to \infty} \nabla f (c_{h(t_n),x_0}(s))= \nabla f (c (s)). $$
The above estimates prove our claim. Using inequality \eqref{estimaflow} together with the dominated convergence theorem, we obtain that
\begin{equation}
    \label{convergenciagradi}
    \lim_{n \to \infty} \int_0^T \left \| \frac{\overline{P}_{t}f (c_{h(t_n),x_0}(s))}{t^2}- \nabla f (c (s)) \right \| ^2 ds = 0.
\end{equation}
On the other hand, since $c_{h(t_n),x_0}(I)$ is solution of Eq.~(\ref{gradientflowequ}), then
$$\begin{array}{rcl} 0 & \ge & \langle \frac{\overline{P}_{t}f (c_{h(t_n),x_0}(s))}{t^2} ,c_{h(t_n),x_0}'(s) \rangle  \\  \,&\ge & \, \langle\frac{\overline{P}_{t}f (c_{h(t_n),x_0}(s))}{t^2}- \nabla f (c (s)) ,c_{h(t_n),x_0}'(s)\rangle +\langle \nabla f (c (s)),c_{h(t_n),x_0}'(s)-c'(s) \rangle\, \\ \, &+ & \langle \nabla f (c (s)),c'(s) \rangle.
\end{array}  $$ 
Using the weak convergence assumption, together with Eq.~(\ref{convergenciagradi}), we conclude that for all points $t_1<t_2$, the following inequality holds
$$0 \ge \int_{t_1}^{t_2}  \langle\nabla f (c (s)),c'(s) \rangle ds= f(c(t_2))-f(c(t_1)).$$
\end{proof}
\label{proposifuerte}
\end{prop}
\textcolor{black}{ The previous result establishes that the flow generated by  $\frac{\overline{P}_{t}f (x)}{t^2}$ approximates a curve $c$ for which the function $f$ is decreasing.}

\section{Algorithm Development} 
\label{applications}
\textcolor{black}{In this section we propose a computational algorithm to approximate the  Riemannian gradient of a function defined on a Riemannian submanifold of the Euclidean space using a set of sample points. We use these approximations as principal directions in gradient-based algorithms as described in Ref.~\cite{sepul}.}  If the function is not differentiable at a point $x$, we say that $x$ is a singularity. Here, we assume that the singularity points form a discrete set.

\par Theorem~\ref{teo1} states that the operator $\overline{P}_{t} f (x)$ can be used to approximate the Riemannian gradient.
An important task  is to compute the integrals involving the operator
$\overline{P}_{t}$, defined in Eq.~(\ref{kerneloperator}). In practical applications, we only have access to  a finite sample  points $x_1, x_2, x_3, \cdots, x_m$  on   $U(x,t^{\delta})$, which are the realizations of i.i.d random variables with probability density function {\it (PDF) } $q$. However, the integral in Eq.~$\eqref{kerneloperator}$ does not depend on the {\it (PDF) } $q$. To address this issue,  for a fixed $x$, we consider the  normalized points
$$ (x_i-x) (f(x_i)-f(x))  \, \,e^{\frac{-\| x_i-x \|^2}{2 t^2}} \, \, / q(x_i),$$
($i=1,\cdots, m$) which are realizations  of i.i.d  random variables  regarding the {\it PDF } $q(x)$. In that case, the Law of Large Numbers {\it LLN } guarantees that
$${\overline{P}_{t} f (x)}= \lim_{m \to \infty} \frac{1}{m \, \,d_t(x)}\sum_{i=1}^m  (x_i-x) \, \,(f(x_i)-f(x))  \, \,e^{\frac{-\| x_i-x \|^2}{2 t^2}} \, \, / q(x_i), $$
where $d_t(x)$ can be computed similarly  using the {\it LLN }
$$ d_t(x)= \lim_{m \to \infty} \frac{1}{m} \sum_{i=1}^m  e^{\frac{-\| x_i-x \|^2}{2 t^2}} \, \, / q(x_i) .$$
\textcolor{black}{The following result establishes a connection between the  tolerance  of the approximation involving the finite sums and the parameters $\delta$, $t$ and $m$. }
\begin{prop} \label{estimaparamefull}
\textcolor{black}{ Let $x$ be a fixed point in $\mathcal{M}$, and $t$ a positive number. Assume that  $q(x)$ is a {\it PDF } on $U(x,t^{\delta})$, and $X_1, X_2, X_3, \cdots, X_m$  are i.i.d multivariate random variables regarding $q$, and that there exists a positive constant $M$ such that 
 $$q(X_i)>M,$$
 for $1 \le i \le m$. Define 
 $$ S^{1}_{m,t}= \frac{1}{m} \sum_{i=1}^m  (X_i-x) \, \,(f(X_i)-f(x))  \, \,e^{\frac{-\| X_i-x \|^2}{2 t^2}} / q(X_i),$$
 and
 $$ S^{2}_{m,t}= \frac{1}{m} \sum_{i=1}^m  e^{\frac{-\| x_i-x \|^2}{2 t^2}} \, \, / q(X_i) .$$
 For a positive constant $C_1$ and $2<u<4 \delta$, we define the set
 $$ A_{t,n}(C_1)= \{\|S^{1}_{m,t} / ( t^{2}S^{2}_{m,t}) - \nabla f(x) \|\le C_1 t^u \}, $$
 where  $n$ and $t$ are the approximation parameters. Thus, there exist  positive constants $C_1$ and $C_2$ such that the probability of the set $A_{t,n}(C_1)$ is bounded below by
 \begin{equation}
 \label{paramecontrodes}
     \mathbb{P}(A_{t,n}(C_1))\ge 1 - \frac{W_4}{( m  e^{-t^{2(\delta-1)}/2} t^{2+d+u})^{2}}.
 \end{equation}    }
 \end{prop}
\begin{proof}
\textcolor{black}{Observe that 
\begin{equation}
    \label{desiguparame1}
\begin{array}{rcl} 
     \|S^{1}_{m,t} / ( t^{2}S^{2}_{m,t}) - \nabla f(x) \| & \le & \|S^{1}_{m,t} / ( t^{2}S^{2}_{m,t}) -{\overline{P}_{t} f (x)}/ {t^{2}}\| + \\ &  & \, \| {\overline{P}_{t} f (x)}/ {t^{2}}- \nabla f(x)\|.
\end{array}
\end{equation}
Since  $\| x_i-x \|<t^{\delta}$, we obtain that
$$ \|S^{2}_{m,t} d_t(x) \| > W_1 e^{-t^{2(\delta-1)}/2},$$
where $W_1$ is a positive constant which does not depend on $t$. In addition, by  Eq. \eqref{estimativanormali}  we have that
$$ \| d_t(x) \| > W_2 t^{d},$$
where $W_2$ is a positive constant. If we define
$$I_{t}=\int_{U(x,t^\delta)} (y-x) (f(y)-f(x))\,e^{\frac{-\| y-x \|^2}{2 t^2}} \, dy,$$
there exists a positive upper bound $W_3$ satisfying $\|I_{t}\| \le W_3$ for all $t$ small enough.
On the other hand,
\begin{equation}
    \label{desipara1}
    \begin{array}{rcl} 
    \|S^{1}_{m,t} / ( t^{2}S^{2}_{m,t}) -{\overline{P}_{t} f (x)}/ {t^{2}}\|  \le  \frac{1}{t^{2}} (  \| S^{1}_{m,t} -I_1 \| / (W_1 e^{-t^{2(\delta-1)}/2}) + \\ \|I_1\| \|d_t(x)-S^{2}_{m,t} \| / (W_2 W_1 t^{d} e^{-2t^{(\delta-1)} }).
    \end{array}
\end{equation}
We define the sets
$$B^{1}_{t,m}(W_1)=\{\| S^{1}_{m,t} -I_1 \| \ge (W_1 e^{-t^{2(\delta-1)}/2}) t^{2+u}\},$$
and 
$$B^{2}_{t,m}(W_2)=\{\| d_t(x)-S^{2}_{m,t} \| \ge (W_2 W_1 t^{d} e^{-t^{2(\delta-1)}/2}) t^{2+u}\}.$$
The Chebyshev's inequality guarantees that
$$\mathbb{P}(B^{1}_{t,m}(W_1) )\le \frac{\sigma_1^2}{m (W_1 e^{-t^{2(\delta-1)}/2}t^{2+u})^{2}},$$
and
$$\mathbb{P}(B^{2}_{t,m}(W_2) )\le \frac{\sigma_2^2}{m (W_2 W_1 t^{d} e^{-t^{2(\delta-1)}/2} t^{2+u} )^{2}},$$
where $\sigma_1^2$ and $\sigma_1^2$ are the respective variance in each case. Therefore,
\begin{equation}
    \label{desigfina}
    \mathbb{P}( B^{1}_{t,m}(W_1)^\complement  \cap  B^{2}_{t,m}(W_2)^\complement)\ge 1 - \frac{W_4}{( m  e^{-t^{2(\delta-1)}/2} t^{2+d+u})^{2}},
\end{equation}
for a proper positive constant $W_4$.
By Theorem \ref{teo1} and Inequalities \eqref{desiguparame1} and \eqref{desipara1}, we have that the following inequality holds
$$   \|S^{1}_{m,t} / ( t^{2}S^{2}_{m,t}) - \nabla f(x) \| \le W_5 t^u,$$
in the set $B^{1}_{t,m}(W_1)^\complement  \cap  B^{2}_{t,m}(W_2)^\complement$, where $W_5$ is a proper positive constant.  The proof is concluded using the previous inequality together with Estimate \eqref{desigfina}. }
\end{proof}      
\textcolor{black}{As a consequence of the fast decay of the exponential function, we obtain the following result:}
\begin{corollary}
    Under the same assumptions of Proposition \ref{estimaparamefull}, we have the inequality
     \begin{equation}
     \mathbb{P}(A_{t,n}(C_1))\ge 1 - W_4\frac{ e^{t^{\delta-1}}}{ m^2}.
 \end{equation}
\end{corollary}
\textcolor{black}{Thus, the convergence rate does not depend on the dimension of the submanifold or the dimension of the ambient space. In this case, convergence is controlled by parameters $t$ and $m$, where $t$ is the approximation parameter and $m$ is the number of sample points.}

In particular,  when the {\it PDF } is the function
\begin{equation}
    \, \, q(y)= e^{\frac{-\| y-x \|^2}{2 t^2}} \, \,/d_t(x),
\label{gausiandistri}
\end{equation}
we can approximate $\overline{P}_{t} f (x)$ using $\mathcal{V}$, where
\begin{equation}
   \mathcal{V}= \frac{1}{m} \sum_{i=1}^m (x_i-x) \, \, (f(x_i)-f(x)).
    \label{estimator}
\end{equation}
This vector is analogous to the weighted gradient operator defined for graphs. See Ref.~\cite{gradientgrafo} for more details. 
\par  \textcolor{black}{Proposition \ref{estimaparamefull} states that once we have chosen the parameters $\delta$ and $t$, the value of $m$ must be greater than $(e^{-t^{2(\delta-1)}/2} t^{2+d+l})^{2}$   to guarantee a proper control in  Inequality \eqref{paramecontrodes}. The parameter $t$  controls how much we approximate the true gradient. Needless to say, a choice of an extremely small $t$ would lead to numerical
instabilities, and thus $t$ in a certain sense would work as a  regularization parameter. In such a scenario, we consider taking the parameter $\delta$ close to $1$ and $t$ moderately small to avoid instabilities generated by selecting the parameter $m$. We shall call $t$ the {\it gradient approximation parameter} and it will be provided as an input to the Algorithm~\ref{algoritgradie}. }

\begin{algorithm}[H]
\begin{flushleft}
\textbf{input}  Sample points $x_1, x_2, x_3 \cdots x_m$ on $U(x,t^{\delta})$ with {\it PDF} $q$, and gradient approximation parameter $t$. \\
\begin{enumerate}
    \item \textbf{for} $i = 1 $ to $m$ \textbf{do}
           \begin{itemize}
               \item $c_i \gets e^{\frac{-\| x_i-x \|^2}{2 t^2}} \, \, / q(x_i)$
           \end{itemize}
\item \textbf{end for}
   \item  $d_{t} \gets \sum_{i=1}^m  c_i $
    \item  $ \mathcal{V} \gets \frac{1}{d_{t}} \sum_{i=1}^m  (x_i-x) \, \,(f(x_i)-f(x)) \, \, c_i $
\end{enumerate}
\textbf{return}   $\mathcal{V}/t^2 $  which  is an approximation for the gradient $\nabla f (x)$ \\
\end{flushleft}
 \caption{ Approximate Gradient Sampling Algorithm}
 \label{algoritgradie}
\end{algorithm}

\textcolor{black}{In Appendix \ref{numericalcomparison}, we explore the numerical consistency of  Proposition \ref{estimaparamefull}, and we also compare the result with the learning gradient approach \cite{muksa}.}



\par We apply Algorithm~\ref{algoritgradie} in a gradient-based optimization method. Intuitively, Proposition~\ref{proposifuerte} says that the energy associated with the gradient decreases along the curve $c$. Therefore,  we can use this curve to find a better approximation for local minimizers, ultimately leading to a derivative-free optimization method. \textcolor{black}{ The proposed algorithm is useful in situations where it is not straightforward to compute the gradient of a function.}
\par {Using Proposition~\ref{proposifuerte}, we have that the flow generated by 
\begin{equation}
    \label{flujoaproximado}
    Dir(x)=\frac{\overline{P}_{t}f (x)}{t^2},
\end{equation}
approximates a curve along which the function $f$   decreases. }
Thus, suggesting that if we use the direction $Dir(x)$ defined in Eq.~(\ref{flujoaproximado}) as the main direction in a gradient-based algorithm, then in a certain way we are approximating the gradient descent method. 
The gradient-based optimization method generated by the direction $Dir(x)$ is described by
$$ x_ {k+1}= \beta_{x_{k}} (x_k - \lambda Dir(x)),$$
where $\lambda$ is some relaxation parameter which defines the step size and $\beta_{x}$ is a local retraction of  $\mathcal{M}$ around the point $x$. 
\par We recall that a local retraction
$\beta_x$ consists of a locally defined
smooth map from a local neighbourhood around $x$ onto the 
manifold $\mathcal{M}$, such that
it coincides with the identity when 
restricted to $\mathcal{M}$. In other words, 
$\beta_X \circ \iota = I_{A}$,
where $A$ is an  open neighbourhood of the point $x$  in the topology induced by $\mathcal{M}$, and $\iota$ is the inclusion map from  $A$  into the ambient space~\footnote{In the framework of  matrix groups or more generally Riemannian submanifolds of $\Rn$ a retraction function is used also in \cite{sepul}.}.
\par The parameter $\lambda$ must be regularly reduced to avoid instabilities in our iteration. We propose to reduce the {\it relaxation parameter} $\lambda$ by a step-scale factor $s_f$ after $l$ consecutive numerical iterations. \textcolor{black}{ This procedure is similar to Armijo point rule described in Ref. \cite{sepul} }. We shall call 
$l$ the {\it sub-iteration control number}.

\par  We update the size  $ \lambda $ of the step such that after a certain number of iterations, it decreases to a pre-conditioned proportion. We do this since the interval for which the curve is defined can be limited, and iterating with a fixed size would generate instabilities in the algorithm. Therefore, if we take smaller step sizes as the number of iterations increases, we obtain better estimates for the minimizer. As the iteration numbers increases, we get closer to a local minimum. For this reason, our stopping criteria is achieved when
$$ |f(x_k)-f(x_{k+1})|\le \epsilon,$$
for a certain tolerance $\epsilon$. The latter will be called the {\it termination tolerance on the function value} and will be provided as an input parameter. Results on the convergence of this algorithm, as well as stopping criteria are described in Ref.~\cite{sepul}.
\par We summarize the above discussion in Algorithm~\ref{algoprinno}.

\begin{algorithm}[H]
\begin{flushleft}
\textbf{input}  Initial guess $x_0$, gradient approximation parameter $t$, relaxation parameter $\lambda$,  sub-iteration control number $l$, termination tolerance $\epsilon$, and step-scale factor $s_f$. \\
\textbf{initialization}\\
 $k \gets 0 $\\
 $counter \gets 0$\\
$x_{min} \gets x_0$ \\
 $x_{-1} \gets x_0$ \\
\textbf{while} \quad $| f(x_{k-1})-f(x_k) | \ge \epsilon$ \textbf{or} $k=0$
\end{flushleft}
\begin{enumerate}
\item $x_{k+1} \gets  \beta_{x_k}( x_k- \lambda \frac{\overline{P}_{t} f (x_k)}{ t^{2}})$
\item \textbf{if} $f(x_{k+1}) < f(x_{min})$ \textbf{do} 
\begin{itemize}
\item $x_{min} \gets x_{k+1}$ 
\end{itemize}
 \item \textbf{end if}
 \item  $k \gets k+1$ 
\item \textbf{if} $l < counter$ \textbf{do} 
\begin{itemize}
\item $counter \gets 0$ 
\item $x_k \gets x_{min}$ 
\item $\lambda \gets \lambda/s_f$
\end{itemize}
\item \textbf{end if}
\item $counter \gets counter+1$
\end{enumerate}
\begin{flushleft}
\textbf{end while}\\
\textbf{return} $x_{min} $  \\
\end{flushleft}
 \caption{Diffusion-map-based  optimization}
 \label{algoprinno}
\end{algorithm}

\subsection{High-dimensional datasets}

\textcolor{black}{
In many optimization problems, the dataset consists of sample points lying in an unknown lower-dimensional submanifold embedded in a high-dimensional space. We propose to use the dimensional reduction method and then,  Algorithm \ref{algoprinno} to solve the optimization problem in the embedded space. This will be done without directly involving the {\it a priori} knowledge of the manifold.}
\par \textcolor{black}{ To be more specific, we assume that the optimization problem under consideration consists on minimizing the cost function $f$ over the dataset $X=\{{x_i} \}_{i=1}^{k}$.  Regarding  the dataset, we suppose that $X \subset \mathcal{M}  \subset\Rn$,  where $n$ is a large number, and $\mathcal{M}$ is a lower-dimensional Riemannian submanifold.
Since the information contains a large number of irrelevant data that make the computing process inefficient, we use the diffusion-maps approach to embed our dataset in a lower-dimensional space. This embedding process allows us to work only with the most important features, and thus, we obtain a better computational performance of the optimization algorithm.  We denote  the embedded  points by
\begin{equation}
   y_i=\psi_{m}^t(x_i),
   \label{embededata}
\end{equation}
where $\psi_{m}^t$  is the  diffusion-map.
We apply Algorithm \ref{algoprinno} to the dataset $Y=\{{y_{i}} \}_{i=1}^{k}$, and the function $\tilde{f}$. Here, the function $\tilde{f}$ is defined as
$\tilde{f}(y_i)=f(x_i),$
for all $x_i \in X$, and $y_i$ the associated point (\ref{embededata}). In this case, we  use the retraction $\beta_x$,  defined as the projection on  $Y$, that is,
$$  \beta_x(z)= \underset{y_i \, \, \in Y}{\arg\min} \, \, \|z-y_i\|.  $$}
\section{Numerical Experiments and Applications}
The following experiments were implemented
 in \MATLAB software, using a desktop computer with the following configuration: Intel  i5 9400 4.1 GHz processor, and 16 GB RAM.

\subsection{Sphere packing problem in dimensions 2 and 3}
\begin{figure}[htp]
\centering
\includegraphics[width=.41\textwidth]{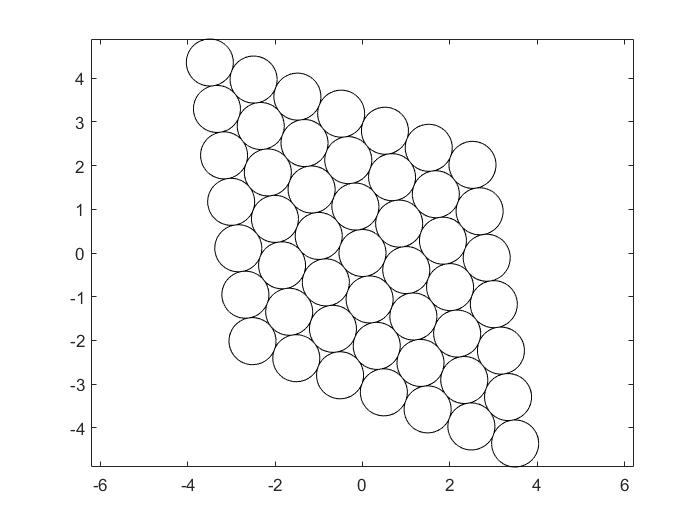}\quad
\includegraphics[width=.41\textwidth]{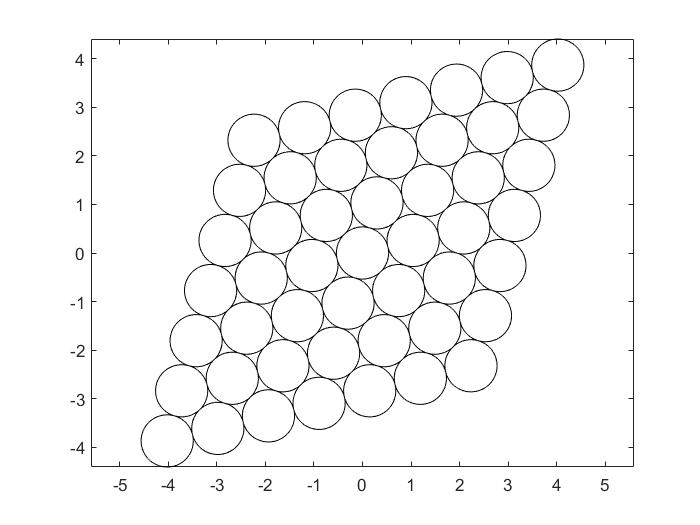}\quad
\includegraphics[width=.41\textwidth]{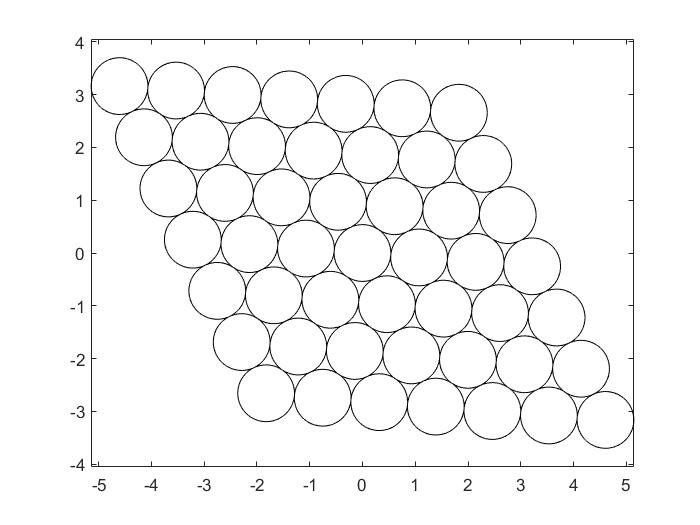}
\medskip
\includegraphics[width=.41\textwidth]{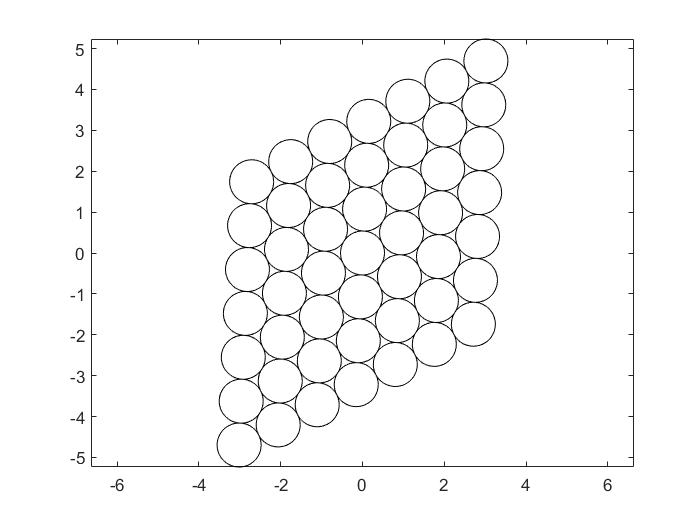}\quad
\includegraphics[width=.41\textwidth]{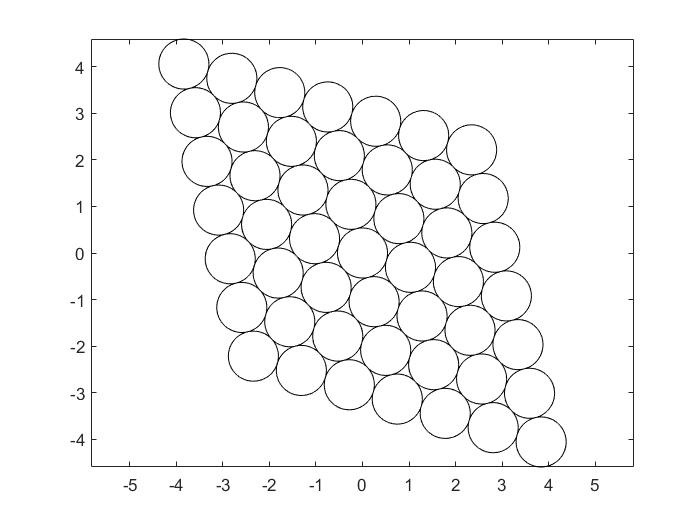}
\caption{Plot of final lattice packing step of five executions to approximate the density $\Delta_{2}$.}
\label{final2d}
\end{figure}

\begin{figure}[htp]
\centering
\includegraphics[width=.41\textwidth]{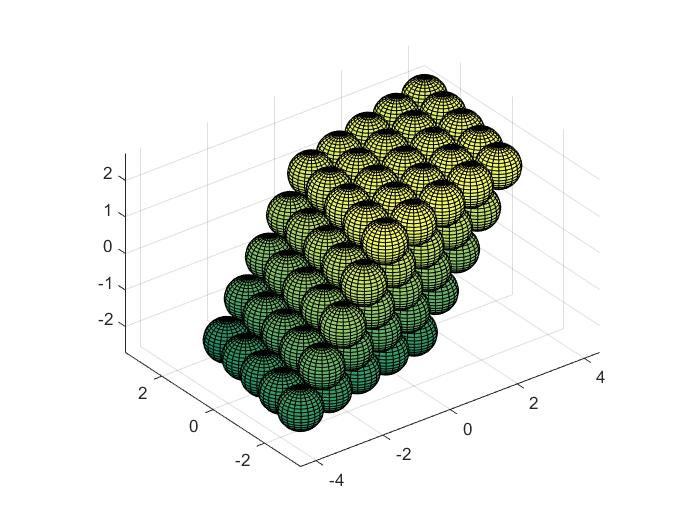}\quad
\includegraphics[width=.41\textwidth]{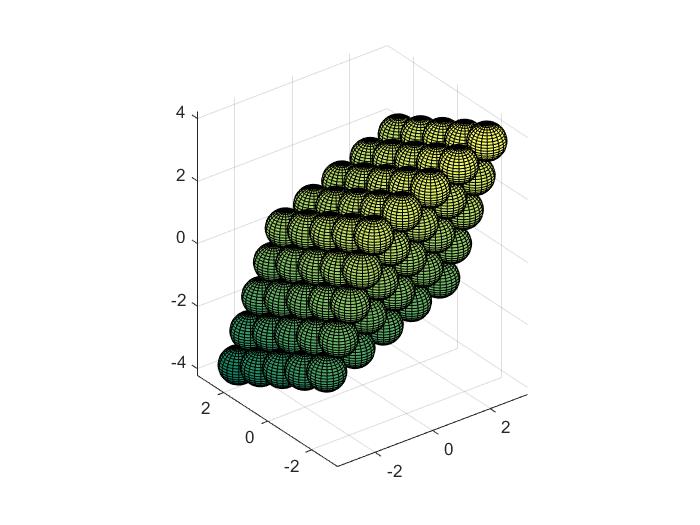}\quad
\includegraphics[width=.41\textwidth]{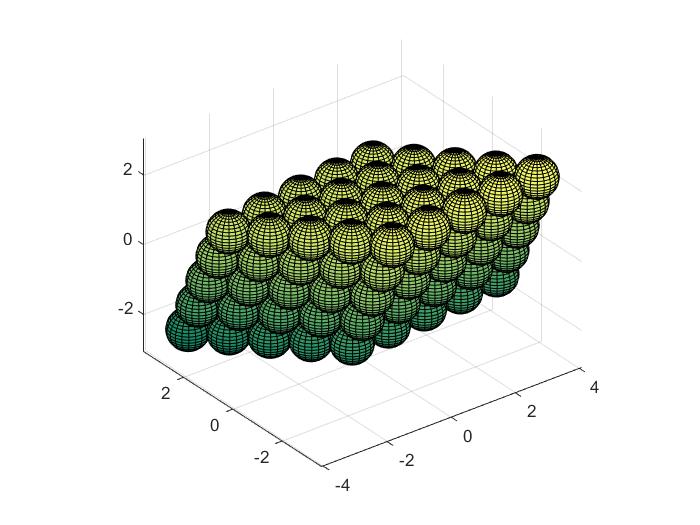}
\medskip
\includegraphics[width=.41\textwidth]{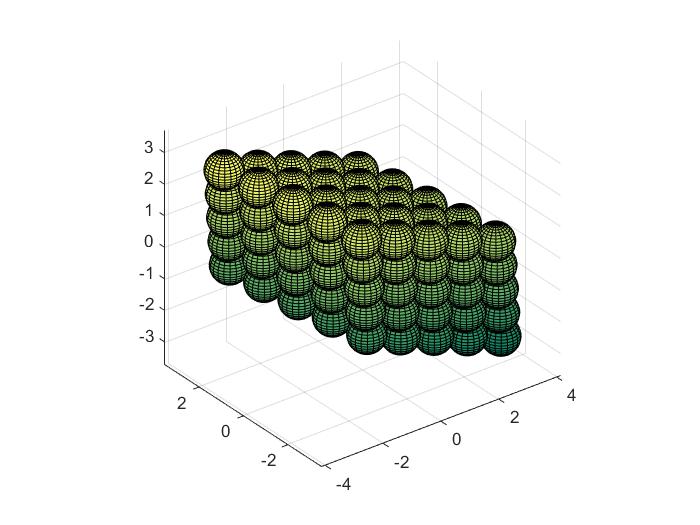}\quad
\includegraphics[width=.41\textwidth]{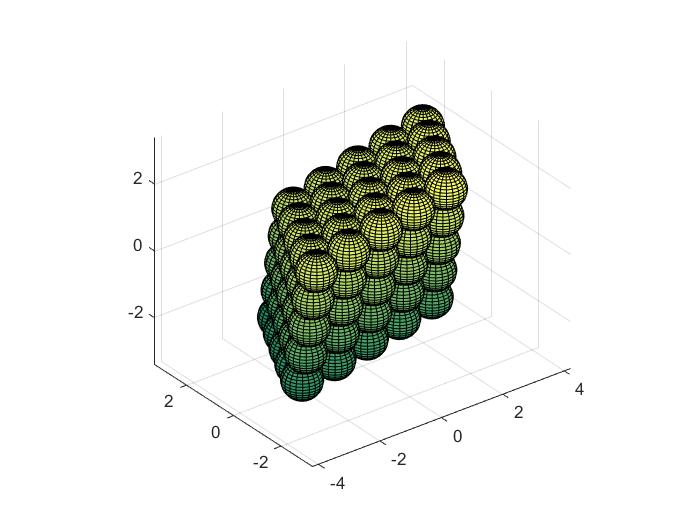}
\caption{Plot of final lattice packing step of five executions to approximate the density $\Delta_{3}.$ }
\label{fianl3d}
\end{figure}

The sphere packing problem \textcolor{black}{in the Euclidean space} poses the following question: How to arrange non-overlapping congruent balls as densely as possible. This problem  has exact solution in dimensions $1, 2, 3, 8$, and $24$. See Refs.~\cite{mari8,mari245}. The one-dimensional sphere
packing problem is the interval packing problem on the line, which is trivial. The two and
three-dimensional cases are far from trivial. In the two-dimensional case the hexagonal packing gives the largest density; see Figure \ref{final2d}.
 The three-dimensional case of packing spheres in $\mathbb{R}^3$ was solved by  Hales in $2005$ and he gave a complex proof, which makes intensive use of computers~\cite{10.2307/20159940}. In this case, the
pyramid arrangement of equally sized spheres filling space is the optimal solution; see Figure \ref{fianl3d}.
 In 2017, Viazovska solved the problem in dimensions eight and twenty-four with coworkers in the latter. See  Refs.~\cite{mari8,mari245}.
\par \textcolor{black}{In this experiment, we reformulate the sphere packing problem as an optimization problem over a manifold, and we use the proposed
methodology to find a computational solution.}
\par We now discuss the problem in more detail. We denote $Vol$ the volume form associated with the Lebesgue measure, and for $x \in \Rn$ and $r$ a positive real number, we denote by $B (x,r)$
 the ball in $\Rn$ with center $x$ and radius $r$.
 \par How do we define a sphere packing in the $n$ dimensional space? To this end, we assume that $C\subset \Rn$ be a discrete set of points such that $2r \le\| x-y \| $, for any two distinct $x,y \in C$, where $r$ is a positive real number. Then, the union
 $$ S=\bigcup_{x\in C} B(x,r),$$
is a sphere packing, and its density $\Delta_{S}$ is defined as
$$\Delta_{S}=\limsup_{r\to \infty} \frac{Vol(S \cap B(0,r))}{Vol \, (B(0,r))}.$$
Intuitively, the density of a sphere packing is the fraction of space covered by the spheres of
the packing. The sphere packing problem consists in knowing what is the supremum  $\Delta_{n}$ over all possible packing densities in $\Rn$. The number $\Delta_{n}$ is called the $n$ dimensional sphere packing constant.
\par One important way to create a sphere packing is to start with a lattice  $\Lambda \subset \Rn$, and center the spheres at the points of $\Lambda$, with radius half the length of the shortest non-zero vectors in $\Lambda$. Such packing is called lattice packing. A more general notion than lattice packing is periodic packing. In periodic packings, the spheres are centered on the points in the
union of finitely many translates of a lattice  $\Lambda$. Not every sphere packing is a lattice packing, and, in all sufficiently large dimensions, there are packings denser than every lattice packing. In contrast, it is proved in Ref.~\cite{Groemer1963} that
periodic packings get arbitrarily close to the greatest packing density. Moreover, in Ref.~\cite{Groemer1963} it is shown that for
every periodic packing $P$ of the form
$$ P=\bigcup_{i=1}^k \bigcup_{x\in \Lambda} (x_i+ B(x,r)), $$
where $\Lambda$  is a lattice, its density is given by
$$ \Delta_{P}=k \frac{Vol \,(B(0,r))}{Vol \, (\Lambda)},$$
where $r=\min_{x,y \in P} \|x-y\|$. 
\par Observe that the density packing is invariant under scaling, that is, for a lattice $\Lambda$ and a positive constant $\alpha$ we have $\Delta_{\alpha \Lambda}=\Delta_{\Lambda}$. Thus, without loss of generality and normalizing if necessary, we can assume that the volume of the lattice is  $Vol \, (\Lambda)=1$. If $b_1, \cdots b_n$ is a basis for $\Lambda$, then our problem can be reformulated as
\begin{equation}
    \begin{alignedat}{3}
&\!\max_{b_1, \cdots b_n}        &\qquad& Vol \, ( B(0,1)) \,  (\frac{g(b_1, \cdots b_n)}{2}) ^n\\
&\text{subject to} &      &det \, (b_1, \cdots, b_n)=1.\\
\end{alignedat}
\label{optipro}
\end{equation}
where $det(\cdot)$ is the determinant function, and the function $g(b_1, \cdots b_n)$ is defined as the minimum  value of $\| z_1\,b_1+ \cdots+ z_n b_n \|_{2}$ over all possible  $(z_1, \cdots z_n) \in \mathbb{Z}^ n \setminus {0}$.
\par Since the function $g$ is defined as a minimum, then this function is non-differentiable at least in the set of orthonormal matrices. In fact, if we consider an orthonormal set $b_1, \cdots, b_n$, then $g(b_1, \cdots, b_n)=1$. In that case, the smooth curve defined as
$$c(t)=(tb_1, \frac{1}{t} b_2, b_3, \cdots,  b_n),$$ 
 for $t>0$, satisfies 
 $$ g(c(t)) = \left\{\begin{array}{lr}
       \frac { 1} {t}, & \text{for }  t \ge 1\\
       t  & \text{for }  t < 1\\
  \end{array} \right. \mbox{. }$$
Since $g(c(t))$ is non-differentiable, then $g$ is not differentiable in $(b_1, \cdots, b_n)$.
\par To apply our approach, we first prove that the function $ g $ is locally Lipschitz.  We write the matrices $A$ and $B$ as the column form $A=[a_1,\cdots a_n]$ and $B=[b_1,\cdots b_n]$, and the special linear group as $SL(n)=\{A \, | \, det(A)=1 \}$. Since the inverse of a  matrix is a continuous function on $SL(n)$, then for $A \in SL(n)$, there exists an open set $ U \ni A $  and a positive constant $D$ such that for all $B \in U$
$$\|B^{-1}\|_{2} \le D. $$
Assume that $g(a_1,\cdots, a_n)=\|A  \,\vec{z} \|_{2}$ and $g(b_1,\cdots, b_n)=\|B  \,\vec{z_2} \|_{2}$ for $\vec{z}, \vec{z_2}  \in \mathbb{Z}^ n \setminus {0}$. In this case $g(b_1,\cdots, b_n)\le \|B  \,\vec{z} \|_{2}$. Then, we have that
\begin{align*}
g(b_1,\cdots, b_n)- g(a_1,\cdots, a_n) & \le \|( A-B) \|_{2}  \,\|\vec{z}\|_{2} \\
&\le \|A^{-1}\|_{2} \|A-B\|_{2} \|A  \,\vec{z}\|_{2}.
\end{align*}
Minkowski's theorem for convex sets \cite{minkowsk} guarantees that for any matrix $A$ with $det(A)=1,$ the estimate $g(A)\le \sqrt{n}$ is satisfied. Thus, we obtain that
$$g(b_1,\cdots, b_n)- g(a_1,\cdots, a_n)\le \sqrt{n}D\|A-B\|_{2}.$$
By symmetry, the above inequality is still valid if we change the order of $A$ and $B$. This proves that $g$ is locally Lipschitz.
\par In dimensions $2$ and $3$ the solutions of the problem in Eq.~ \eqref{optipro} are  $\Delta_{2}=\frac{\pi }{2 \sqrt{3}}$ and  $\Delta_{3}=\frac{\pi}{3 \sqrt{2}}$, respectively. In these dimensions the maximizers are the hexagonal lattice, Figure \ref{final2d}, and the pyramid lattice packing, Figure \ref{fianl3d}.
\par \textcolor{black} { Observe that the  problem in Eq.~\eqref{optipro} can be considered as an optimization problem on the manifold $SL(n)$. We use our approach to find the maximizers in dimensions $2$ and $3$. Since maximizing the function $g$ is equivalent to minimizing $-g$, then we apply our approach to the function $-g$.
We use Algorithm \ref{algoprinno} to minimize the function $-g$, and thus Algorithm \ref{algoritgradie} to compute $\overline{P}_{t}f (x)$. In this experiment, we use the {\it PDF} function $q$ defined as in Eq~(\ref{gausiandistri}) to compute the gradient. In this case, the approximation  is given by Eq.~\eqref{estimator}. We generate a total of $m=20$ sample points  from the normal distribution for the parameter $\delta=0.99$ using the \MATLAB function {\it normrnd}, and then projected to the manifold $SL(n)$ using the retraction given by
\begin{equation}
    \beta_A(b_1,\cdots, b_n)=\frac{\left(\mathrm{sign}(\det(B)\right) b_1,b_2,\cdots, b_n)}{|\det(B)|^{\frac{1}{n}}}.
    \label{retracspehere}
\end{equation}
Since $\Delta_n\le 1$, then, we take a small initial step size to get a better performance of our methodology. Our initial guess $x_0$, is the identity matrix and initial parameters $t=10^{-5}, \,\, \lambda=0.1,  \, \, l=10,  \, \,\epsilon=10^{-10}, \, \, s_f=1.1$.
We note that these are the parameters for which we obtain better results.}
\par We use  the Exhaustive Enumeration Algorithm proposed in Ref.~\cite{schnorr1994lattice} to compute the function $g$. The implementation of this algorithm is provided in the GitHub repository~\cite{softwa}  using \MATLAB. 
In Figures \ref{final2d} and \ref{fianl3d}, we plot the final step of each execution of the proposed algorithm in dimensions $2$ and $3$. Observe that in all executions, the final step approximates the optimal sphere packing illustrated in Figures~\ref{final2d} and \ref{fianl3d} in each dimension (to rotations). This fact was verified by calculating the error as shown in Figure~\ref{erroresdel}.

\begin{figure}[htp] 
    \centering
    \subfloat[Approximation error for $\Delta_{2}$ using Algorithm~\ref{algoprinno}]{%
        \includegraphics[width=0.5\textwidth]{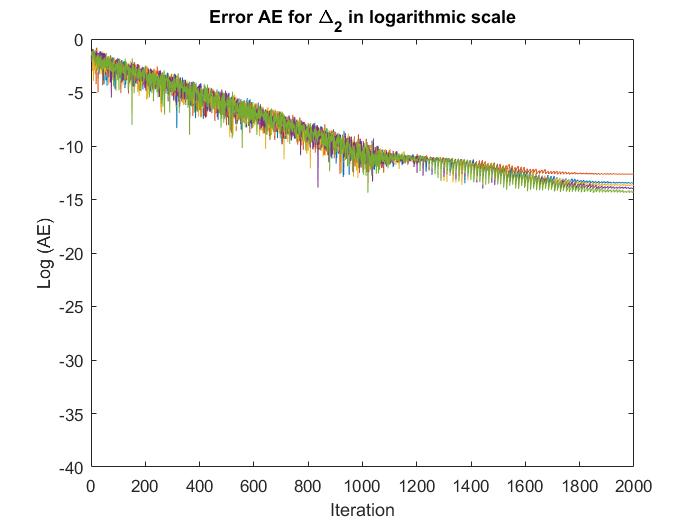}%
        }%
    \subfloat[Approximation error for $\Delta_{3}$ using Algorithm~\ref{algoprinno}]{%
        \includegraphics[width=0.5\textwidth]{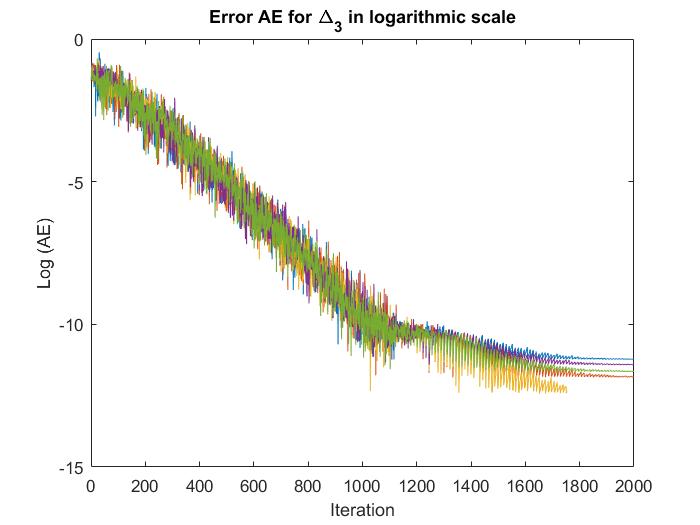}%
        
 }%
 \\
 
     \subfloat[Approximation error for $\Delta_{2}$ using PSO]{%
        \includegraphics[width=0.5\textwidth]{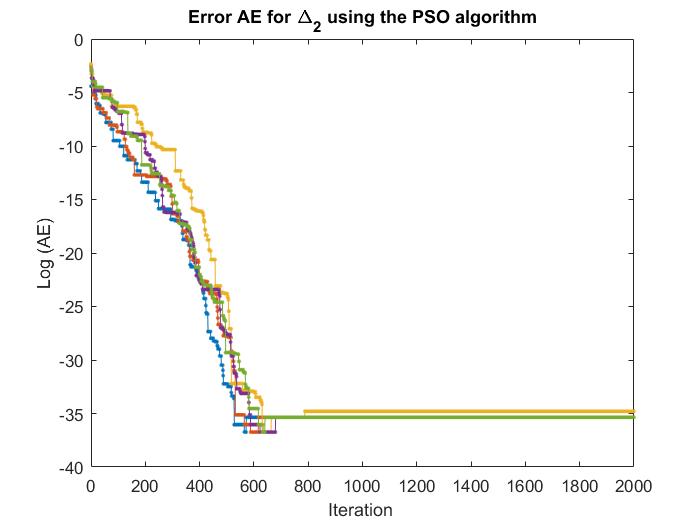}%
        }%
    \subfloat[Approximation error for $\Delta_{3}$ using PSO]{%
        \includegraphics[width=0.5\textwidth]{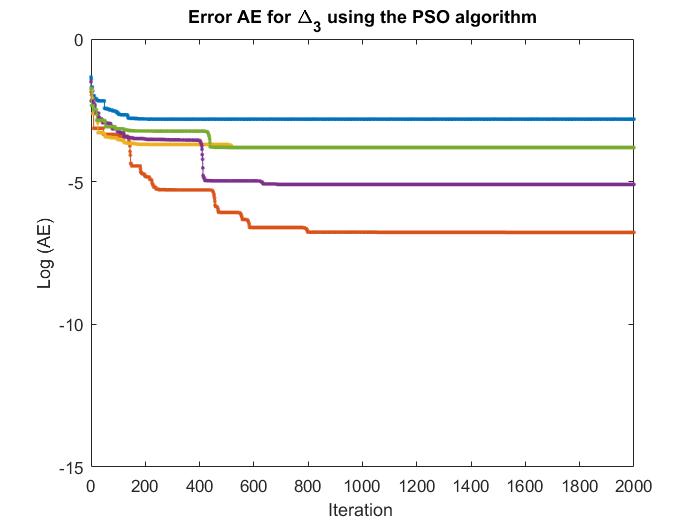}
        
 }%
 \\
     \subfloat[Approximation error for $\Delta_{2}$ using Nelder-Mead]{%
        \includegraphics[width=0.5\textwidth]{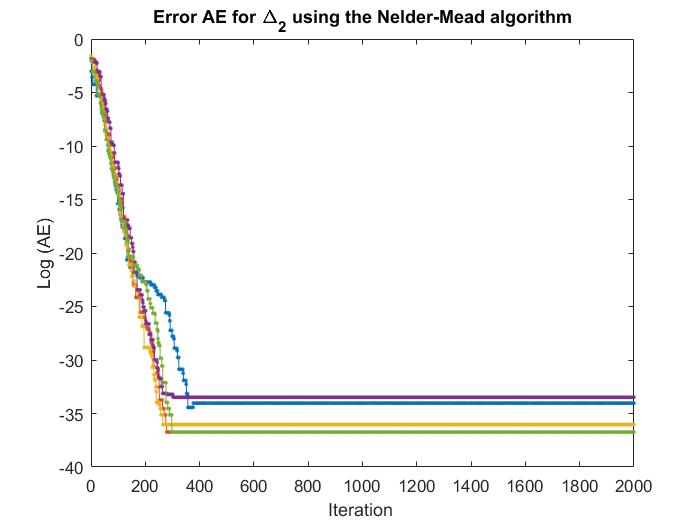}%
        }%
    \subfloat[Approximation error for $\Delta_{3}$ using Nelder-Mead]{%
        \includegraphics[width=0.5\textwidth]{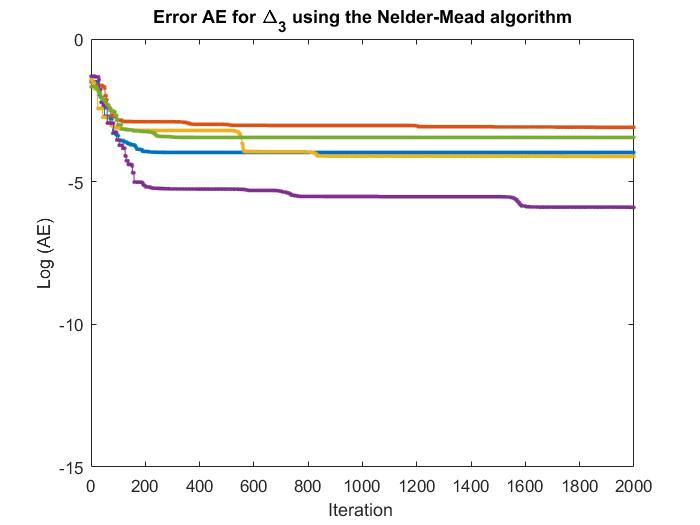}} 
        
  \caption{Plot of the absolute error ($AE$)  generated by five executions using Algorithm~\ref{algoprinno}, PSO and Nelder-Mead. Here, we use the logarithmic scale.}
    \label{erroresdel}
\end{figure}

\par \textcolor{black}{ We now compare the proposed algorithm with the \textsc{PSO} and \textsc{Nelder-Mead}, for that, we run five different executions for the different algorithms. In Figure~\ref{erroresdel}, we plot  the absolute error $(AE)$ of approximating $\Delta_{2}$ and $\Delta_{3}$ for the iteration value $x_n$. Each color represents a different execution. The \textsc{PSO} and \textsc{Nelder-Mead} algorithms are implemented in the \textsc{Manopt toolbox} using default parameters. We implement the PSO algorithm with $40$ particles. }\\
\textcolor{black}{In addition, we test the proposed method to compute an approximation for the densities $\Delta_4$ and $\Delta_5$ using the previous setting. Although the problem remains unsolved in these dimensions, the best packing densities in the literature \cite{conway_a._bannai_1993,cohn2022} are    $0.6168502750680849 \cdots$ for $n=4$ and $0.4652576133092586…$ for $n=5$. Through the execution of the different algorithms, for both cases, the best packing density is obtained using the proposed methodology. In fact, the Algorithm~\ref{algoprinno} for the case $n=4$, gives an optimal packing density equal to $0.616825892885318$ and for $n=5$, gives $0.465218060094373$.}
\textcolor{black}{Thus, as evidence, we observe that the proposed methodology outperforms the $PSO$ and $Nelder-Mead$ free derivative algorithms, for dimensions greater than $2$.}
\textcolor{black}{We emphasize that the proposed methodology focuses on cases where in each iteration the only information available is a set of sample points lying in an unknown manifold. In such case, the solvers \textsc{PSO} and \textsc{Nelder-Mead} cannot be executed.}

\begin{figure}[htp] 
    \centering
    \subfloat[Optimal packing density $\Delta_{4}$ using Algorithm~\ref{algoprinno}]{%
        \includegraphics[width=0.5\textwidth]{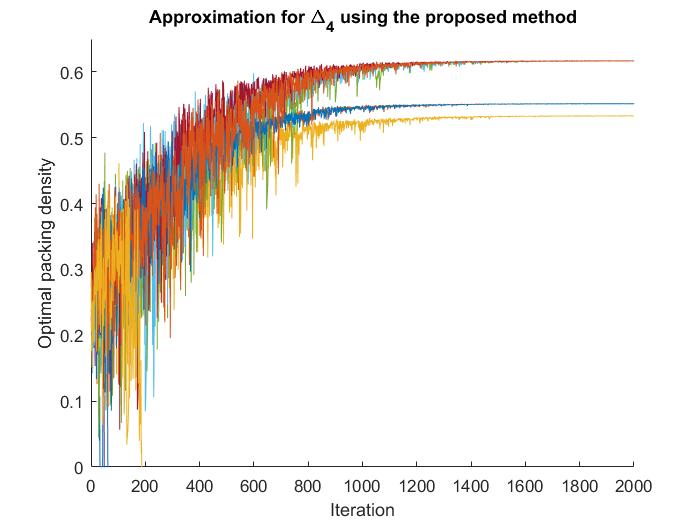}%
        }%
    \subfloat[Optimal packing density  $\Delta_{5}$ using Algorithm~\ref{algoprinno}]{%
        \includegraphics[width=0.5\textwidth]{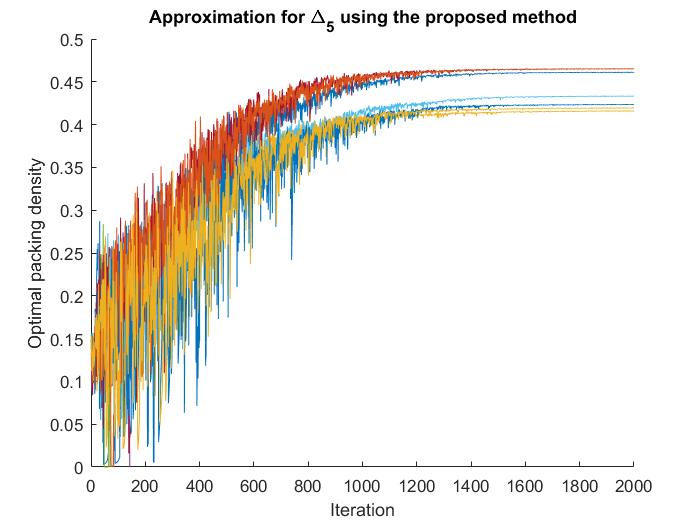}%
        
 }%
 \\
 
     \subfloat[Optimal packing density  $\Delta_{4}$ using PSO]{%
        \includegraphics[width=0.5\textwidth]{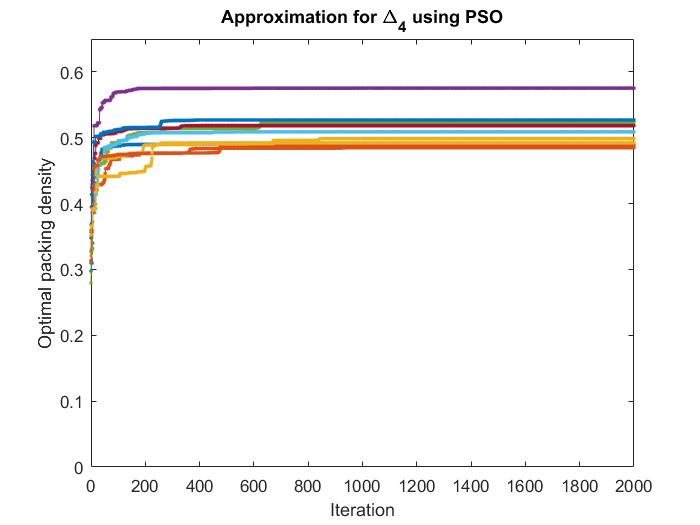}%
        }%
    \subfloat[Optimal packing density  $\Delta_{5}$ using PSO]{%
        \includegraphics[width=0.5\textwidth]{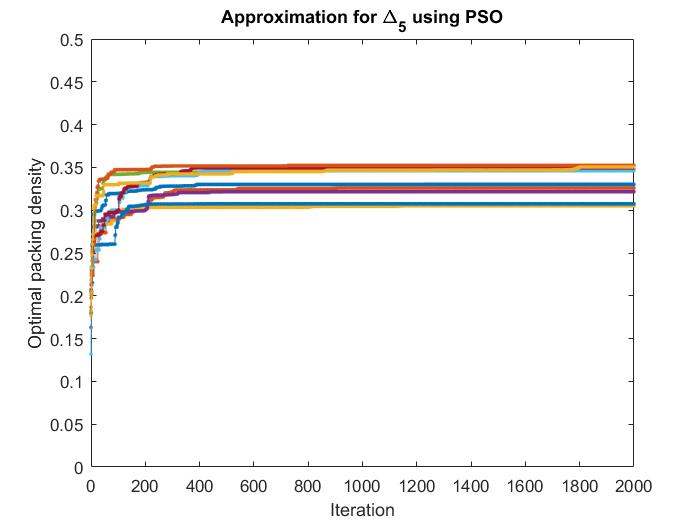}
        
 }%
 \\
     \subfloat[Optimal packing density  $\Delta_{4}$ using Nelder-Mead]{%
        \includegraphics[width=0.5\textwidth]{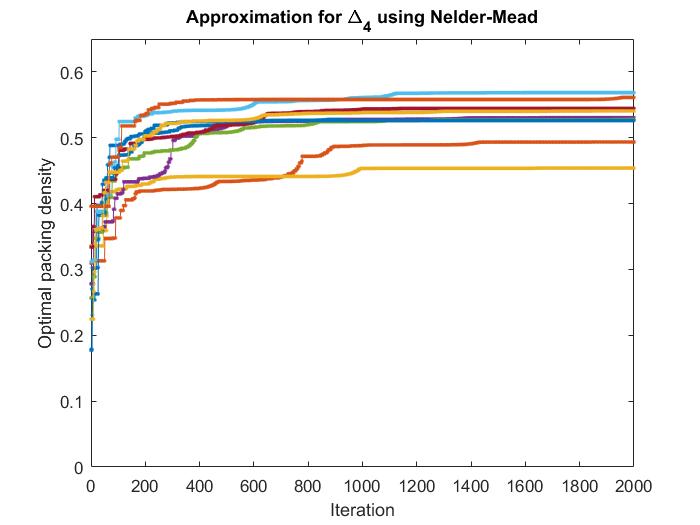}%
        }%
    \subfloat[Optimal packing density  $\Delta_{5}$ using Nelder-Mead]{%
        \includegraphics[width=0.5\textwidth]{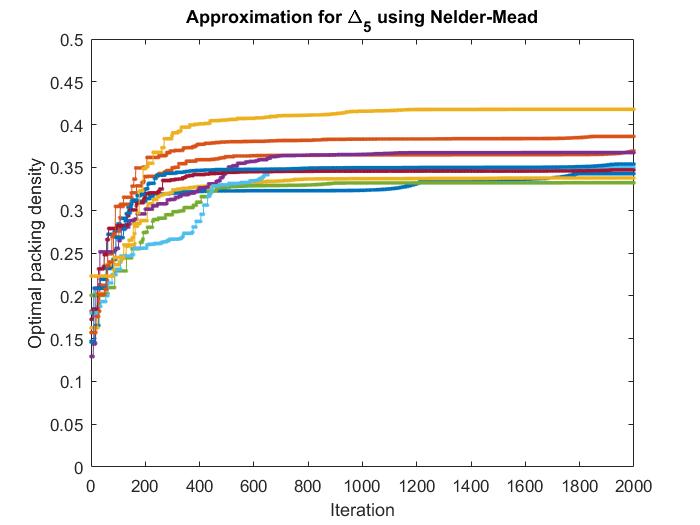}} 
        
  \caption{Plot of the best packing density generated by ten executions using Algorithm~\ref{algoprinno}, PSO, and Nelder-Mead.}
    \label{erroresdel}
\end{figure}


\subsection{ Tomographic reconstruction from  unknown random angles}
\textcolor{black}{ Tomographic reconstruction is a widely studied problem in the field of inverse problems. Its  goal  is to reconstruct an object from its angular projections. This problem has many applications  in  medicine, optics and other areas. We refer the reader to Refs.~\cite{radon3,radon1,radon2,natterer} for more details .}
\par \textcolor{black}{Classical reconstruction methods are based on the fact that the angular position is known. See Ref.~\cite{radon3}. In contrast, there are many cases for which the angles of the projections are not available, for instance,  when the object is moving. The latter is a nonlinear inverse problem, which  can be  more difficult when compared to the classical linear inverse problem.}
\par \textcolor{black} {Now, we explain the problem in more details. Suppose that $f: \R^2 \to \R_{\ge 0} $ describes the density of an object, and let $\theta$ be an angle. We define the one-dimensional tomographic projection  over the angle $\theta$ as
$$\mathbb{P}_\theta f (x)=\int f(R_{\theta}(x,y)) \, dy,$$
where $R_{\theta}(x,y)$ is the counterclockwise rotation of the two-dimensional vector $(x,y)$ with  respect to the angle $\theta$.  Since $$ \int |\mathbb{P}_{\theta_i}f(x) |  \, dx =\int f(x,y) \, dy  dx,$$
thus, normalizing if necessary, we also assume  that $\|\mathbb{P}_{\theta_i}f \|_{L^1}=1$.}
\textcolor{black}{ The problem under consideration consists in reconstructing the density $ f $ with the knowledge of  projections $\mathbb{P}_{\theta_1} f, \mathbb{P}_{\theta_2} f, \cdots \mathbb{P}_{\theta_k} f$, where the angles $ \theta_1, \theta_2, \cdots \theta_k $ are unknown. If through some method the rotations are known, then  we can obtain the density function $ f $ using classical reconstruction methods.}
\par
\begin{figure}[htp] 
    \centering
\tikzset{every picture/.style={line width=0.75pt}} 
\begin{tikzpicture}[x=0.75pt,y=0.75pt,yscale=-1,xscale=1]

\draw    (230.8,76.8) .. controls (326.6,20.6) and (478.8,120.8) .. (399.8,150.8) ;
\draw    (230.8,76.8) .. controls (125.8,149.8) and (249.8,179.8) .. (399.8,150.8) ;
\draw    (325.8,159.8) -- (365.09,139.28) -- (415.8,112.8) ;
\draw    (276.8,159.8) -- (325.09,136.28) -- (405.8,96.8) ;
\draw    (233.8,154.8) -- (273.09,134.28) -- (381.8,82.8) ;
\draw    (202.8,140.8) -- (242.09,120.28) -- (352.8,68.8) ;
\draw    (192.8,116.8) -- (232.09,96.28) -- (307.8,60.8) ;
\draw    (456,211) .. controls (481.28,210.71) and (482.75,178.05) .. (451.74,176.74) ;
\draw [shift={(449.8,176.7)}, rotate = 360] [color={rgb, 255:red, 0; green, 0; blue, 0 }  ][line width=0.75]    (10.93,-3.29) .. controls (6.95,-1.4) and (3.31,-0.3) .. (0,0) .. controls (3.31,0.3) and (6.95,1.4) .. (10.93,3.29)   ;
\draw    (75.8,215.8) -- (500.81,168.92) ;
\draw [shift={(502.8,168.7)}, rotate = 173.71] [color={rgb, 255:red, 0; green, 0; blue, 0 }  ][line width=0.75]    (10.93,-3.29) .. controls (6.95,-1.4) and (3.31,-0.3) .. (0,0) .. controls (3.31,0.3) and (6.95,1.4) .. (10.93,3.29)   ;
\draw  (27.8,217.1) -- (505.8,217.1)(75.6,48.8) -- (75.6,235.8) (498.8,212.1) -- (505.8,217.1) -- (498.8,222.1) (70.6,55.8) -- (75.6,48.8) -- (80.6,55.8)  ;
\draw [line width=3]  [dash pattern={on 7.88pt off 4.5pt}]  (195,58) -- (234.3,191.25) ;
\draw [shift={(236,197)}, rotate = 253.57] [fill={rgb, 255:red, 0; green, 0; blue, 0 }  ][line width=0.08]  [draw opacity=0] (16.97,-8.15) -- (0,0) -- (16.97,8.15) -- cycle    ;
\draw [line width=3]  [dash pattern={on 7.88pt off 4.5pt}]  (309,47) -- (348.3,180.25) ;
\draw [shift={(350,186)}, rotate = 253.57] [fill={rgb, 255:red, 0; green, 0; blue, 0 }  ][line width=0.08]  [draw opacity=0] (16.97,-8.15) -- (0,0) -- (16.97,8.15) -- cycle    ;
\draw [line width=3]  [dash pattern={on 7.88pt off 4.5pt}]  (257,54) -- (296.3,187.25) ;
\draw [shift={(298,193)}, rotate = 253.57] [fill={rgb, 255:red, 0; green, 0; blue, 0 }  ][line width=0.08]  [draw opacity=0] (16.97,-8.15) -- (0,0) -- (16.97,8.15) -- cycle    ;
\draw [line width=3]  [dash pattern={on 7.88pt off 4.5pt}]  (357,42) -- (396.3,175.25) ;
\draw [shift={(398,181)}, rotate = 253.57] [fill={rgb, 255:red, 0; green, 0; blue, 0 }  ][line width=0.08]  [draw opacity=0] (16.97,-8.15) -- (0,0) -- (16.97,8.15) -- cycle    ;
\draw    (472.8,70) -- (403.75,85.56) ;
\draw [shift={(401.8,86)}, rotate = 347.3] [color={rgb, 255:red, 0; green, 0; blue, 0 }  ][line width=0.75]    (10.93,-3.29) .. controls (6.95,-1.4) and (3.31,-0.3) .. (0,0) .. controls (3.31,0.3) and (6.95,1.4) .. (10.93,3.29)   ;

\draw (492,233.4) node [anchor=north west][inner sep=0.75pt]    {$x$};
\draw (48,53.4) node [anchor=north west][inner sep=0.75pt]    {$y$};
\draw (120,111.4) node [anchor=north west][inner sep=0.75pt]    {$f( x,y)$};
\draw (483,188.4) node [anchor=north west][inner sep=0.75pt]    {$\theta_i $};
\draw (241,199.4) node [anchor=north west][inner sep=0.75pt]    {$x_{1}$};
\draw (296,194.4) node [anchor=north west][inner sep=0.75pt]    {$x_{2}$};
\draw (350,191.4) node [anchor=north west][inner sep=0.75pt]    {$x_{3}$};
\draw (398,187.4) node [anchor=north west][inner sep=0.75pt]    {$x_{4}$};
\draw (462,44.4) node [anchor=north west][inner sep=0.75pt]    {$Object$};
\draw (233,13.4) node [anchor=north west][inner sep=0.75pt]    {$X-rays$};
\end{tikzpicture}
\caption{Tomography  of an object.}
\label{figutomogrejem}
\end{figure}
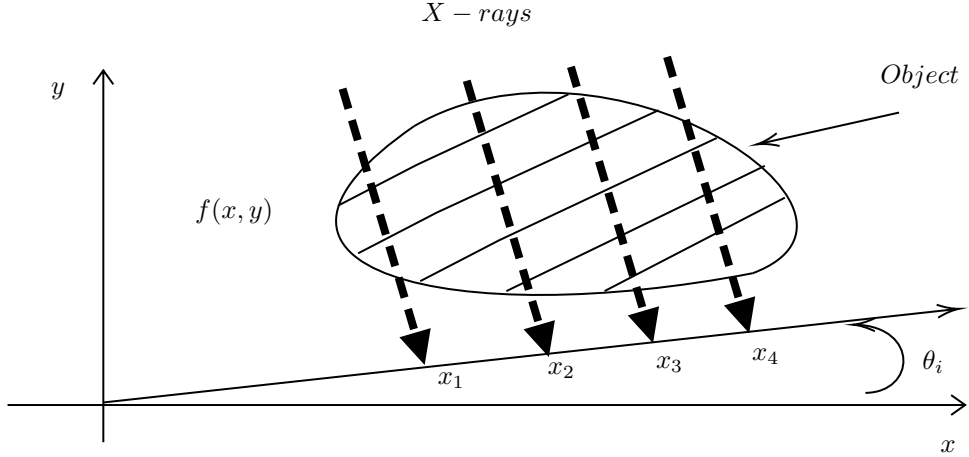

 \textcolor{black}{In Ref. \cite{angucoif} an approach using the graph Laplacian is proposed to deal with this problem. However, the  difficulty in using the previous approach is that it assumes  {\it a  priori} the knowledge of the distribution of the angles $\{ \theta_i \}_{i=1}^k$}. That is, it is necessary to assume the Euclidean distance between two consecutive angles. We use our methodology to tackle the latter problem, the road-map of our approach is established in Algorithm \ref{algorittomogra}. Let $DS$ be the dataset defined as  the set of all tomographic projections
\begin{equation}
    \label{dataset}
    DS= \{\mathbb{P}_{\theta_i} f \}_{i=1}^k.
\end{equation}
If we assume that the density function $f$ has compact support, then a straightforward computation gives
\begin{align}
 \int \mathbb{P}_{\theta_i}f (x) \, x \, dx & =\int \int \langle(x,y) , (f(R_{\theta}(x,y),0)) \rangle dx dy  \nonumber \\
 & = \int \int \langle (x,y) ,  R_{\theta}(f(x,y) , 0)) \rangle dx dy  \nonumber\\
 & = \langle \tilde{V} , R_{\theta_i} \,( 1 , 0 ) \rangle,
 \label{equacionrelacionrado}
\end{align}
where $\tilde{V}$ is the two-dimensional vector 
$$\tilde{V}=(\int \int x f(x,y) \, dx dy , \int \int y f(x,y) \, dx dy ).$$
For practical purposes, we consider the discretization of the projection $\mathbb{P}_{\theta_i}f $ as the multidimensional vector given by 
$$\overline{\mathbb{P}_{\theta_i}}f= ( \mathbb{P}_{\theta_i} f (x_1) , \mathbb{P}_{\theta_i} f (x_2) , \cdots , \mathbb{P}_{\theta_i} f (x_l) ),$$ 
 where $x_1 < x_2 < \cdots < x_l $ are equally spaced fixed  points on the $x$ axis that describe the projection onto the angle $\theta_i$. See Figure~\ref{figutomogrejem}.
\par  Let $X$ be  the  multidimensional vector
$$X=(x_1, x_2 , \cdots , x_l).$$
The discretization of the integrals in Eq.~\eqref{equacionrelacionrado} gives 
\begin{equation}
    \frac{1}{h} \, \langle  \,\overline{\mathbb{P}_{\theta_i}} f , X \rangle \approx  \langle \tilde{V} , R_{\theta_i} \,( 1 , 0 ) \rangle ,
    \label{eqprinang}
\end{equation}
where $ h $ is the distance between two consecutive points. Equation~\eqref{eqprinang} allows to estimate, except for a possible sign and translation, the angle $\theta_i$. Namely, if the two-dimensional vector $\tilde{V}$ has angle $\tilde{\theta}$, then, we recover $\theta_i$  using the expression
\begin{equation}
\cos \, (\theta_i-\tilde{\theta}) \approx \frac{1}{ h \, \,\|\tilde{V} \|_{2} } \langle \overline{\mathbb{P}_{\theta_i}} f , X \rangle .
    \label{cosenang}
\end{equation}
In this case, we use Eq.~(\ref{eqprinang}) to compute the value $\|\tilde{V} \|$ as
\begin{equation}
\|\tilde{V} \|_{2} \approx \max_{\theta_i} \left | \frac{1}{ h  } \langle \overline{\mathbb{P}_{\theta_i}} f , X \rangle  \right |. 
    \label{angulomaxi}
\end{equation}
We remark that  in this approach we do not compute the two-dimensional vector $\tilde{V}$, instead, we compute the norm $\|\tilde{V} \| $ using Eq.~\eqref{angulomaxi}. Observe that to solve the optimization Problem in Eq.~~\eqref{angulomaxi} it is sufficient to assume that ${\theta_i} \in [0, \pi]$.
\par Once we solve the previous optimization problem, we  use Eq.~(\ref{eqprinang}) to calculate the angle $\theta_i-\tilde{\theta}$. Observe that if we do not determine the sign of the $\theta_i-\tilde{\theta}$, then a flipping effect appears on the reconstructed object, resulting in an image with many artifacts.  We apply our gradient estimates to determine the sign of the angle.  For  that, we assume that  the angles are  distributed on the interval  $  I=[0, \pi]$, and consider the numbers
\begin{equation}
m_1= \min_{i} |\theta_{i}-\tilde{\theta}| \quad \quad and \quad \quad M_1=\max_{i} |\theta_{i}-\tilde{\theta}|.
    \label{angupeque}
\end{equation}
Since the maximum of the optimization problem in Eq.~\eqref{angulomaxi} is reached for some $\theta_i$, then $m_{1}=0$ or $M_{1}=\pi$. Without loss of generality, it is enough to consider the case $m_1=0$. In fact, if $M_{1}=\pi$,  then we reflect the angles over the $y$-axis. Furthermore, changing the order if necessary we assume that
\begin{equation}
    0=|\theta_{1}-\tilde{\theta}| < |\theta_{2}-\tilde{\theta}|< \cdots < |\theta_{k}-\tilde{\theta}|.
    \label{anguorga}
\end{equation}
\par \textcolor{black}{ We observe that our dataset $(DS)$ defined as in Eq.~(\ref{dataset}) lies in the curve $c(I)$, which is  parameterized by 
$$ c(\theta)=\mathbb{P}_{\theta}f,$$
and in our case this parametrization is unknown. The main idea in our algorithm is to use the  gradient flow of the function $g$ on the manifold $c(I)$, where $g:c(I) \to \R $ is defined as
\begin{equation}
g(Y)= \frac{1}{ h  } \langle Y, X \rangle.
    \label{levantamiento1}
\end{equation}
The importance of the gradient flow in our method lies in the fact that in a local neighborhood of the vector associated with the angle $0$, the gradient flow divides the dataset into two different clusters that determine the sign of the associated angles. This fact is proved using the approximation \eqref{cosenang} and the fact that the derivative of  $\it{cosine}$ is an odd function on the real line. }
\par \textcolor{black}{ Before  initializing  our algorithm we divide the indices $\tilde{A}=\{i\}_{i=1}^k$ as follows. We select a fixed number $s$, which represents the size of the partition, and we consider the decomposition  $k=us+r$, where $u$ and $r$ are non-negative integers with $r<s$. Then, we define the sets
\begin{equation}
\tilde{A}_i=\{is+1, is+2, \cdots, (i+1)s\},
    \label{indice1}
\end{equation}
for $i \in \{0, 1, 2, \cdots,q-1\}$, and 
\begin{equation}
\tilde{A}_q=\tilde{A} \, \, \setminus \bigcup_{i=0}^{u-1} \tilde{A}_i.
    \label{indice2}
\end{equation}
We use the partition $\{\tilde{A}_i\}_{i=1}^q$ to represent the local geometry of the dataset. For that, we consider the subset $DS_i$ of $DS$, defined as
\begin{equation}
DS_i=\{\mathbb{P}_{\theta_j}f \, | j \in A_i\}.
    \label{dataparti}
\end{equation}
The first step in our algorithm is to determine the sign of angles in a local neighborhood of $0$, for that, we use the diffusion-map algorithm to embed the dataset $\overline{DS}_1=DS_1 \cup DS_2 \cup DS_3$ into the two-dimensional space $\R^2$. We endow this embedded dataset with the counting measure.  Once  the dataset is embedded,  we proceed to compute the  approximation for $\overline{P}_{1} \tilde{g}$ as described in Algorithm \ref{algoritgradie}. Here, we select  the points $x_1, x_2, x_3 \cdots x_m$ as  the $m$  closest points to $x$.
Since we only are interested in the direction induced by the gradient, then we propose to  reduce the computational cost of the execution using the approximation
\begin{equation}
\overline{\mathcal{V}}= \sum_{i=1}^m  (x_i-x) \, \,(\tilde{g}(x_i)-\tilde{g}(x))  \, \,e^{\frac{-\| x_i-x \|^2}{2 }},
    \label{normaestima}
\end{equation}
where, the function $\tilde{g}$ is  such that for each two-dimensional embedded point $y \in \R^2$ associated with vector $Y \in DS $, the value of  $\tilde{g} (x)$ is defined as 
\begin{equation}
    \tilde{g}(y)=g(Y).
    \label{levantamiento2}
\end{equation}
The two-dimensional representation of the dataset allows determining the sign of the angles $\theta_{i}-\tilde{\theta}$ regarding the orientation of the flow generated by the function $\tilde{g}(y)$. This is done by observing that locally the set of gradient vectors associated with positive angles and the set of gradient vectors associated with negative angles are separated by a hyperplane.  Since $\theta_{2}-\tilde{\theta}$  is the smallest nonzero angle, then we use its gradient to define a hyperplane that separates the sets mentioned above. To be more specific, we separate the sets according to the sign of the inner product of its gradient with the gradient associated with $\theta_{2}-\tilde{\theta}$. We remark that in the first step we only classify the sign of angles associated with points lying in $DS_1 \cup DS_2$, to avoid instabilities generated by computing the gradient of the boundary points lying in $DS_3$.}
\par \textcolor{black}{The second step is to proceed inductively to determine the sign of the remaining angles as follows. Assume that for $2\le i$ the sign of the angles associated with points lying in the set $DS_{i}$ is determined, and consider the dataset $\overline{DS}_i=DS_i \cup DS_{i+1}$. As in the first step, we use diffusion-maps to embed this dataset into $\R^2$. Observe that the function $g$ has not critical points on $\overline{DS}_i$. Then, the two-dimensional representation is divided at most into two clusters, for which each cluster represents the set of points with the same sign. We determine the sign of each cluster according to the sign of angles associated with points in $DS_{i}$ lying in the corresponding cluster. For practical purposes, we define the sign of each angle $\theta_{i}-\tilde{\theta}$ as the sign of the angle previously determined with the closest two-dimensional representation.  We run this step until all the signs are determined. We summarize this reconstruction method in Algorithm \ref{algorittomogra}. We remark that the choice of parameters $s$ and $m$  have to  be modestly small to avoid instabilities in our algorithm.}

\begin{algorithm}[]
\begin{flushleft}
\textbf{input} Tomographic projections $DS= \{Y_i\}_{i=1}^k$, where $Y_i=\mathbb{P}_{\theta_i} f$, size of the partition $s$. \\ 
\begin{enumerate}
\item Normalize the dataset $DS$ such that $\|\mathbb{P}_{\theta_i}f \|_{L^1}=1$ for all $i$.
\item Compute $\|\tilde{V} \|_{2} $ solving the optimization problem~\ref{angulomaxi}.
\item  Determine the angles $\theta_{i}-\tilde{\theta}$ using Eq.~(\ref{cosenang}).
\item Compute $M_1$  as in Eq.~(\ref{angupeque}).
\item If $M_1=\pi$,  then we proceed to reflect the angles $\tilde{\theta_i}$ over the $y$-axis.
\item Construct $DS_i$ following Eqs~(\ref{indice1}),~(\ref{indice2}), and~(\ref{dataparti}).
\item Use the diffusion-map approach to embed the dataset $DS_1 \cup DS_2 \cup DS_3$  into $\R^2$.
\item Compute  $\overline{P}_{1} \tilde{g}$  using the approximation~(\ref{normaestima}), where $\tilde{g}$ is defined  in Eqs.~(\ref{levantamiento1}) and~(\ref{levantamiento2}).
\item Determine the sign of the angles $\tilde{\theta_i}$ associated with points in $DS_1 \cup DS_2 $, according to the sign of the inner product of the associated gradient with the gradient associated with $\theta_2$.
    \item \textbf{for} $j = 2 $ to $s$ \textbf{do}
           \begin{itemize}
               \item Use the diffusion-map approach to embed the dataset $DS_j \cup DS_{j+1}$  into $\R^2$.
               \item   Determine  the sign of each angle $ \tilde{\theta_i}$  in $DS_{j+1}$ as the sign of angle previously determined with the closest two-dimensional representation.
           \end{itemize}
\item \textbf{end for}
 \item  Reconstruct the  signed angles.
\end{enumerate}
\end{flushleft}
 \caption{ Tomographic reconstruction from unknown random angles}
 \label{algorittomogra}
\end{algorithm}

The computational complexity of all the embeddings is $O(us^3)$, which corresponds to the complexity of the eigenvalue decomposition. On the other hand, the complexity of all gradient computations is $O(s)$, and the computational complexity of the other procedures described in  Algorithm \ref{algorittomogra} is $O(s)$. Thus, Algorithm \ref{algorittomogra} runs with a $O(us^3)$ complexity which improves the $O(u^3s^3)$ complexity of the algorithm proposed in Ref.~\cite{angucoif}.

\par
\begin{figure}
    \centering
    \subfloat[Shepp–Logan phantom]{%
        \includegraphics[width=0.7\textwidth]{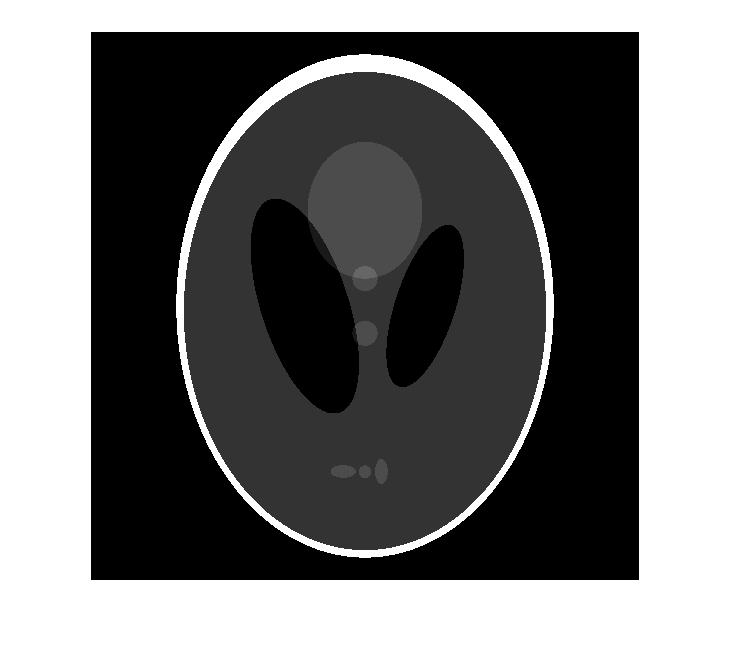}%
        }%
    \hfill%
    \subfloat[Sample image of a knee]{%
        \includegraphics[width=0.5\textwidth]{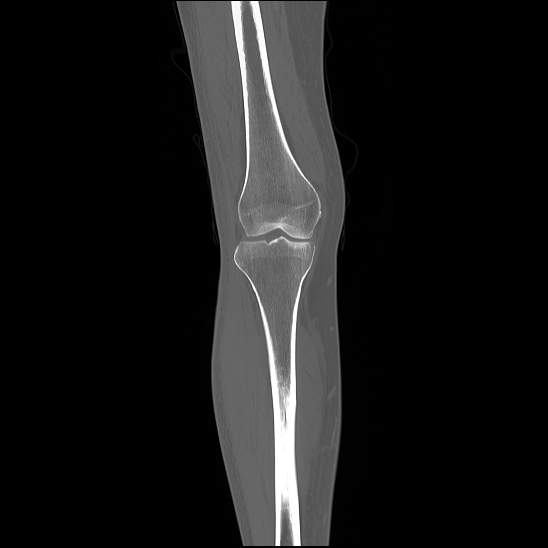}%
       
        }%

    \caption{Picture of the Shepp–Logan phantom (a), and a knee sample image (b). Source for the latter Ref.~\cite{radiografi}.}
    \label{figtomogr}
\end{figure}

  \textcolor{black}{We test our algorithm on the tomographic reconstruction of two objects. The first is the  Shepp–Logan phantom,  and the second is a computed tomography of  a knee taken from Ref.~\cite{radiografi}. See Figure  \ref{figtomogr}. In this experiment, we generate $k=2 \times 10^3$  random  points uniformly distributed in $[0, \pi]$. The parameters used in Algorithm \ref{algorittomogra} are $s=20$, and $m=10$. The tomographic projections $\mathbb{P}_{\theta_1} f, \mathbb{P}_{\theta_2} f, \cdots \mathbb{P}_{\theta_k} f$ are computed using \MATLAB`s {\texttt {radon} }  function.  We add random noise to these projections, for that, we consider the dataset of the form
\begin{equation}
    \label{tomografialevelerro}
    \mathbb{P}_{R_i}^{\varepsilon} f=\mathbb{P}_{R_i} f+ \eta W, 
\end{equation}
where $W$ is a white noise. Our purpose is to recover the density $f$, using only the measurements $\mathbb{P}_{R_i}^{\varepsilon} f$, regardless of their respective angles. }
\par
\begin{figure}
    \centering
            \subfloat[Shepp–Logan phantom]{%
            \includegraphics[width=0.4\textwidth]{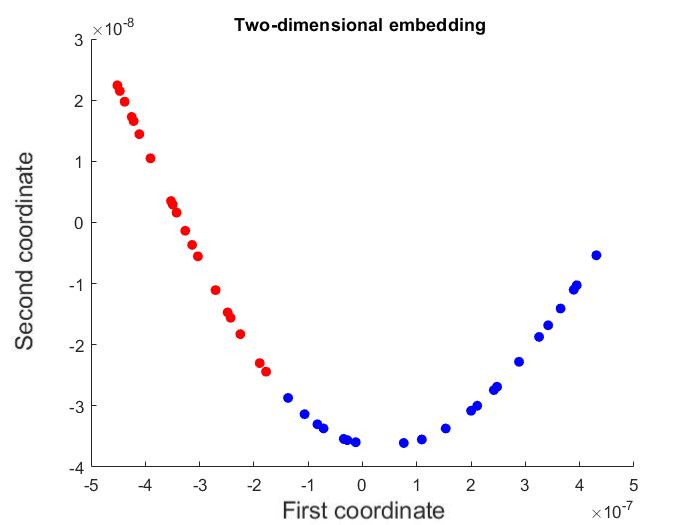}%
        \includegraphics[width=0.4\textwidth]{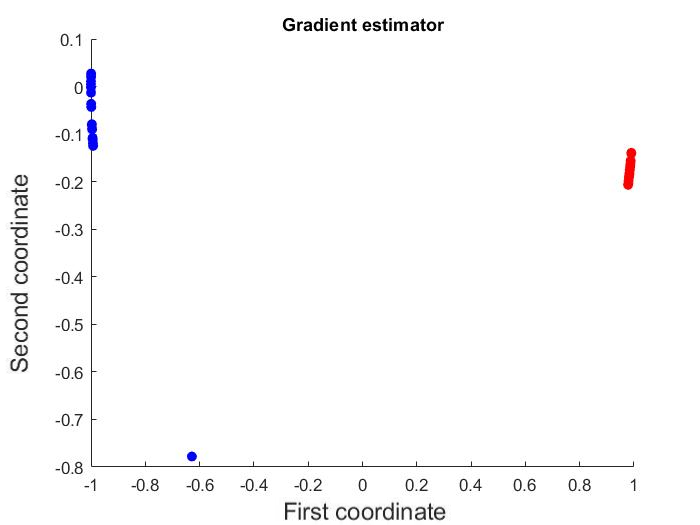}%
  }%
 \medskip
 
  \subfloat[Sample image of the knee]{%
            \includegraphics[width=0.4\textwidth]{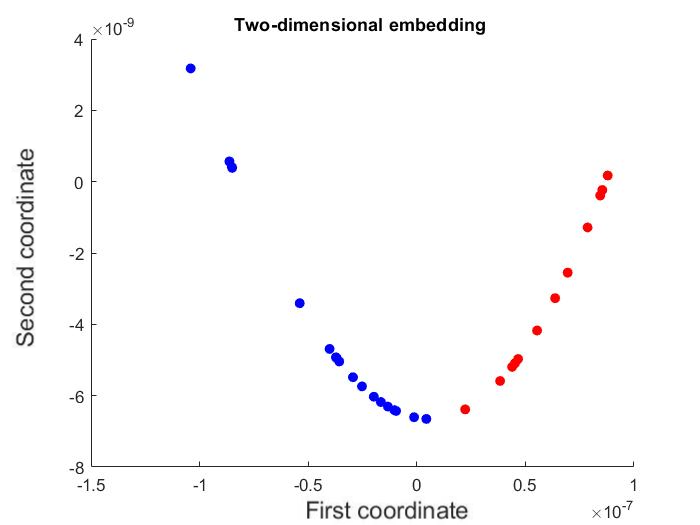}%
        \includegraphics[width=0.4\textwidth]{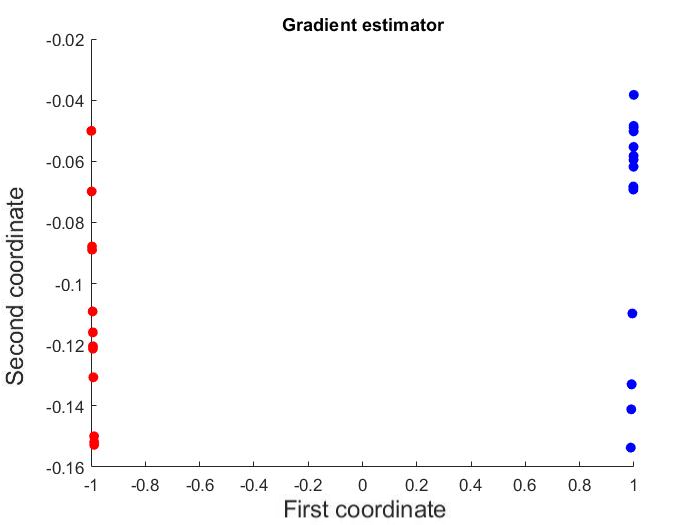}%

        }%
        \medskip

 \caption{Plot of the first two-dimensional embedding (left), and their associated gradient approximation (right). In this experiment, the angle sample is uniformly distributed on $[0,\pi]$. Each color represents a different sign. Figure (a) corresponds to the Shepp–Logan phantom, and Figure (b) to the image of the knee. }
    \label{twoembedd}
\end{figure}

\begin{figure} 
    \centering
            \subfloat[Shepp–Logan phantom]{%
            \includegraphics[width=0.4\textwidth]{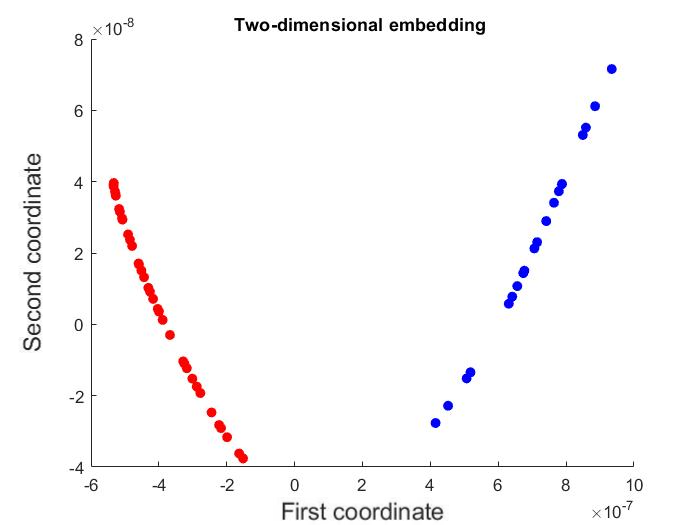}%
            }
             \subfloat[Shepp–Logan phantom]{%
             \includegraphics[width=0.4\textwidth]{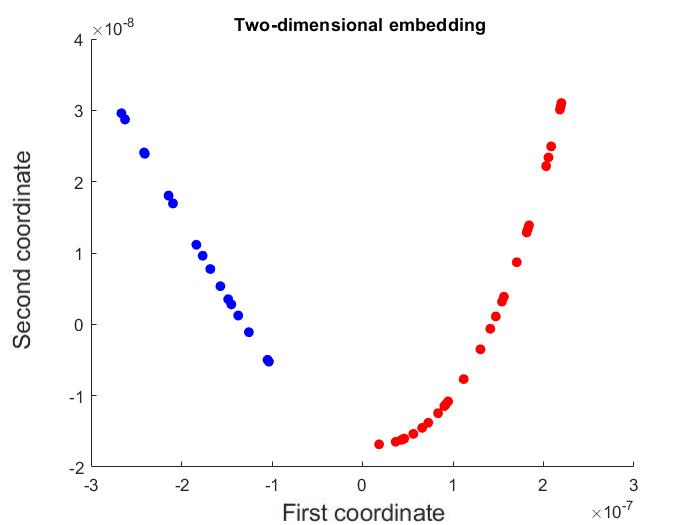}%
      
  }%
 \medskip

 \caption{Plot of the second two-dimensional embedding (left). Figure (a) corresponds to the Shepp–Logan phantom, and Figure (b) to the image of the knee. }
    \label{ste}
\end{figure}

  \textcolor{black}{To illustrate how  Algorithm \ref{algorittomogra} works, we plot the two essential steps in the method. In Figure \ref{twoembedd}, we plot the first two-dimensional embedding and their respective gradient approximation defined in Eq.~(\ref{normaestima}). Points with blue color are associated with positive angles and those with red color with negative angles. Furthermore, in Figure \ref{ste} , we plot the second two-dimensional embedding of our method. We observe that our method performs effectively in dividing the dataset into two different clusters according to the sign of the corresponding angle.}
\par
\begin{figure}
    \centering
            \subfloat[$\eta=0$]{%
            \includegraphics[width=0.4\textwidth]{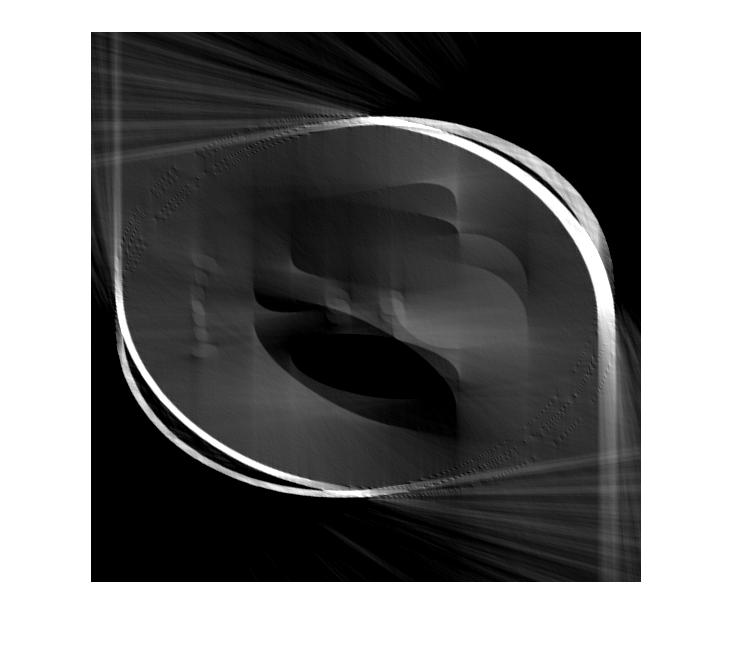}%
        \includegraphics[width=0.4\textwidth]{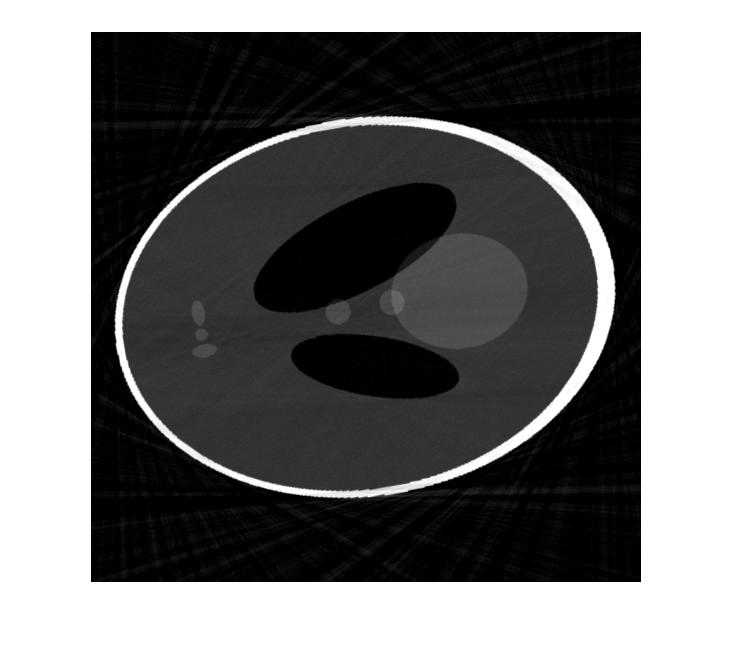}%
  }%
 \medskip
 
            \subfloat[$\eta=0.05$]{%
            \includegraphics[width=0.4\textwidth]{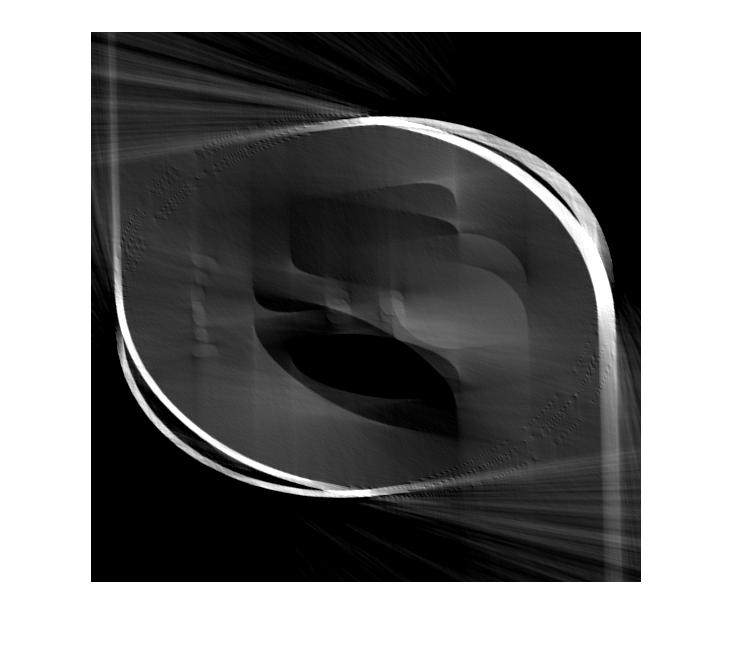}%
        \includegraphics[width=0.4\textwidth]{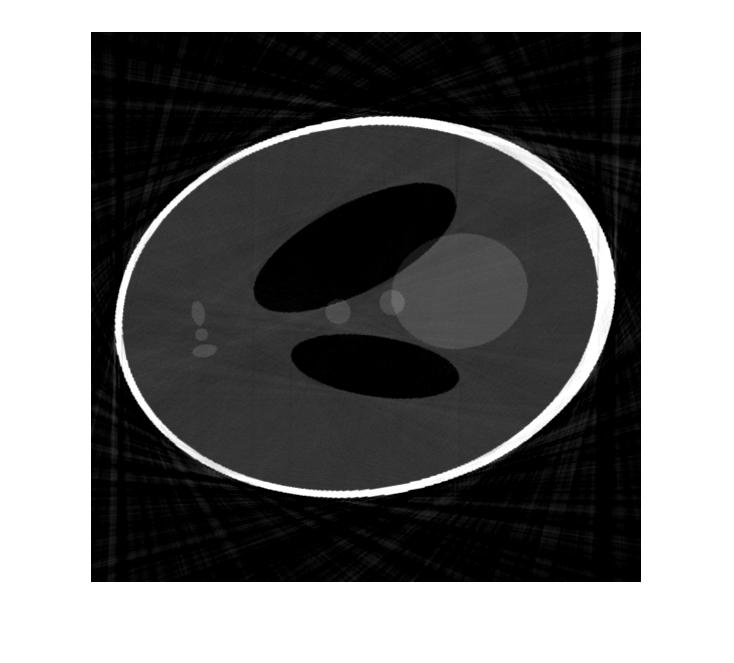}%

        }%
        \medskip
        
                    \subfloat[$\eta=0.1$]{%
            \includegraphics[width=0.4\textwidth]{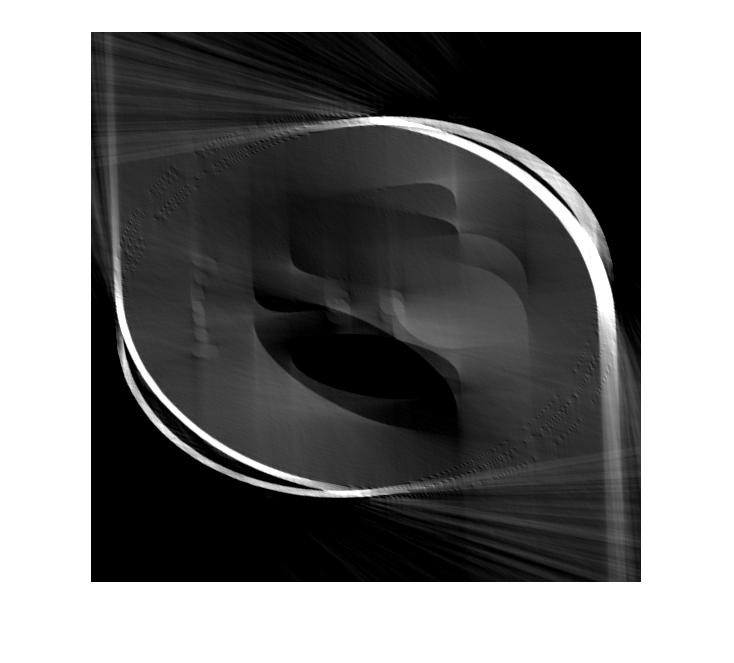}%
        \includegraphics[width=0.4\textwidth]{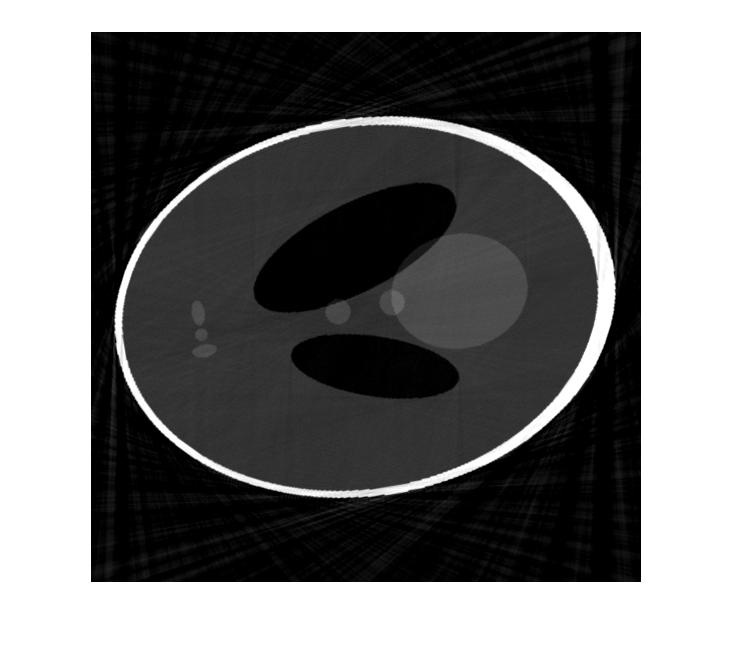}%

        }%

 \caption{Reconstructed  Shepp–Logan phantom for several additive errors $\eta$  as in Eq.~\eqref{tomografialevelerro}.
    The images on the left are obtained without determining the sign of each angle, and the images on the right are obtained by implementing our algorithm. }
    \label{reconstrufan}
\end{figure}

\begin{figure}
    \centering
            \subfloat[$\eta=0$]{%
            \includegraphics[width=0.4\textwidth]{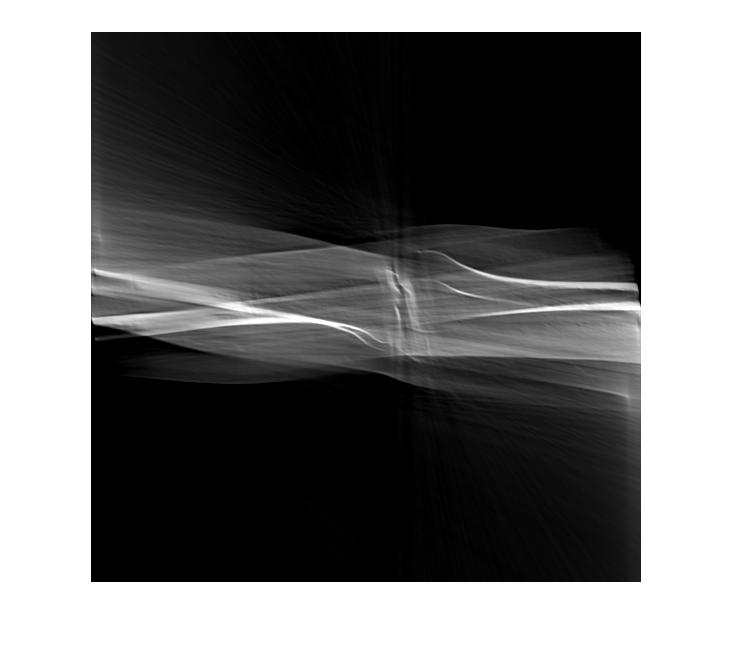}%
        \includegraphics[width=0.4\textwidth]{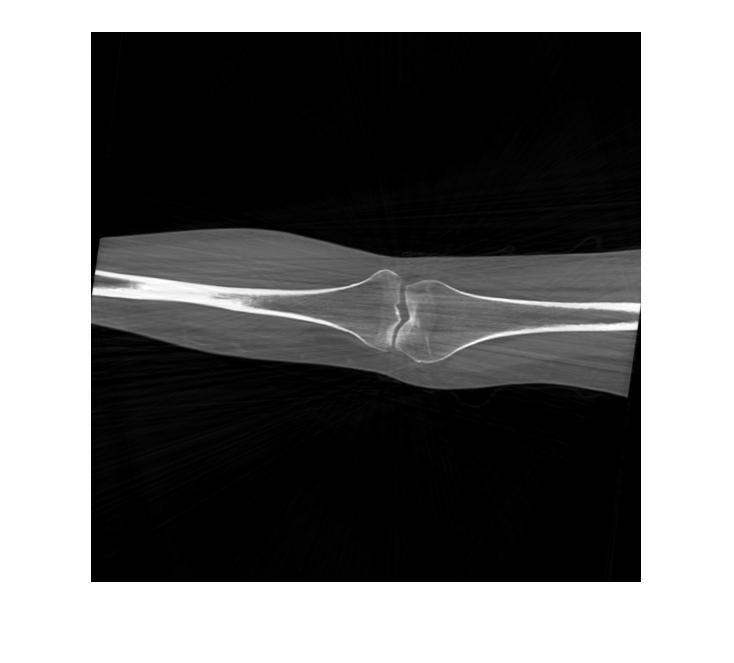}%
  }%
 \medskip
 
            \subfloat[$\eta=0.05$]{%
            \includegraphics[width=0.4\textwidth]{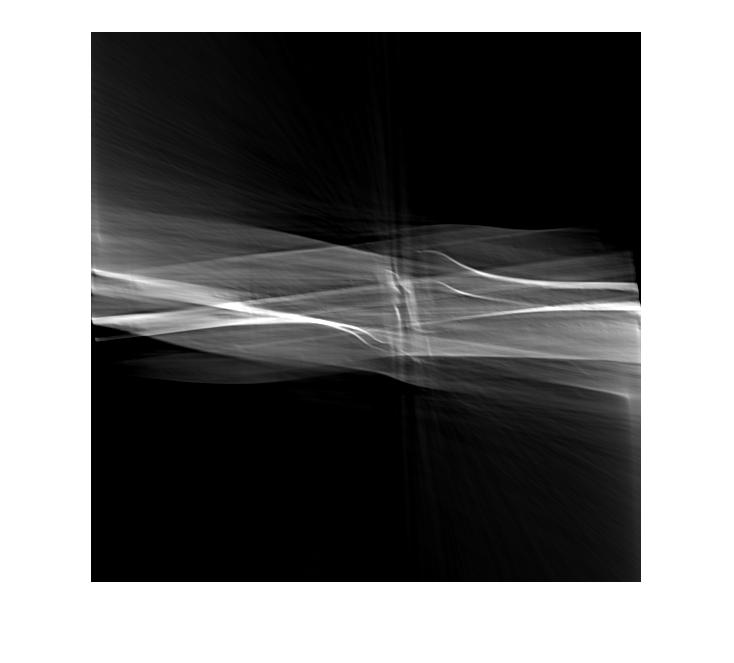}%
        \includegraphics[width=0.4\textwidth]{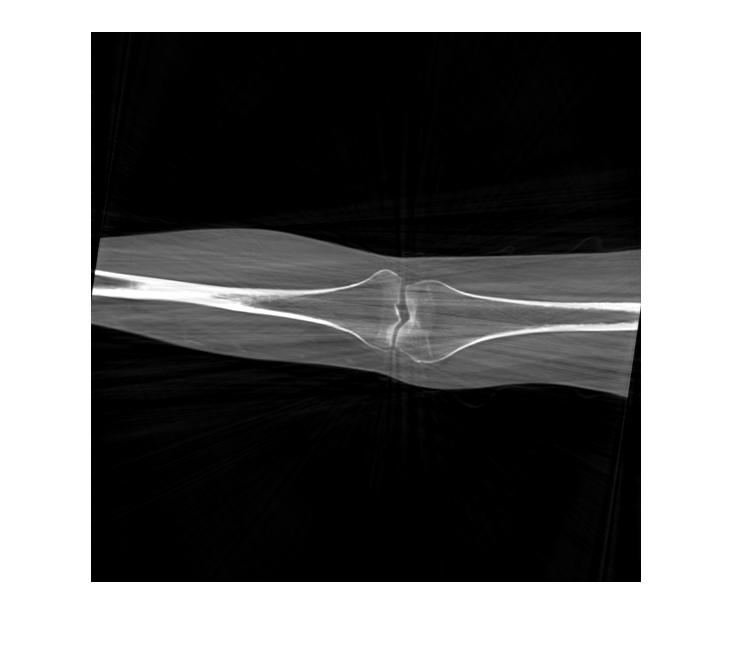}%

        }%
        \medskip
        
                    \subfloat[$\eta=0.1$]{%
            \includegraphics[width=0.4\textwidth]{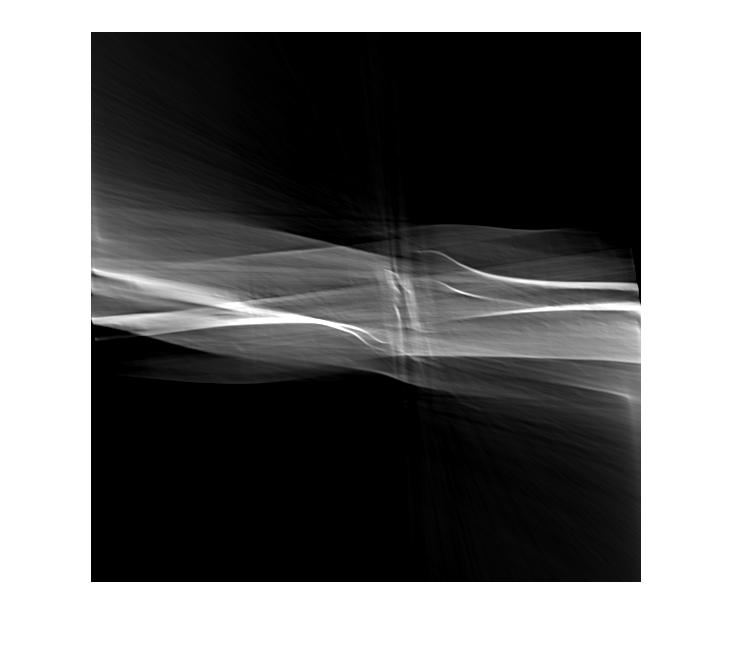}%
        \includegraphics[width=0.4\textwidth]{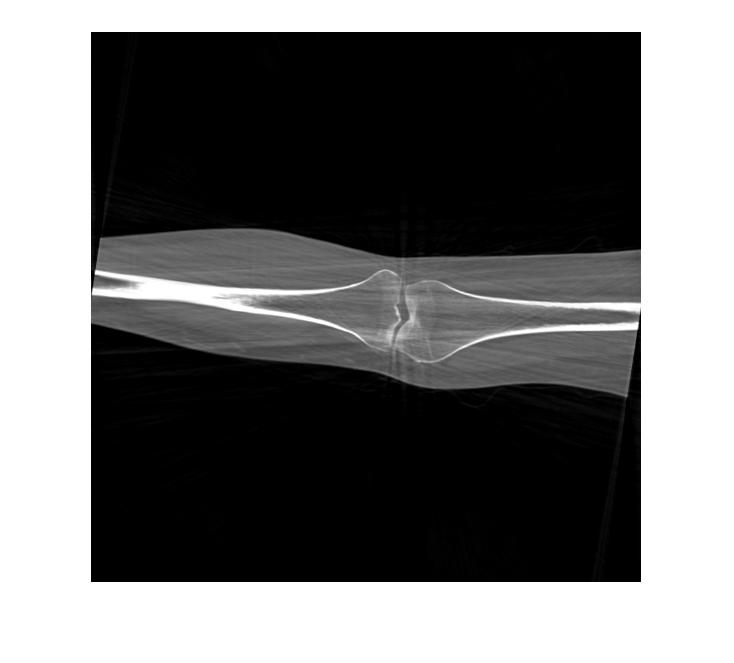}%

        }%

 \caption{Reconstructed knee tomography for several additive errors $\eta$  as in Eq.~\eqref{tomografialevelerro}.
    The images on the left are obtained without determining the sign of each angle, and the images on the right are obtained by implementing our algorithm. }
    \label{reconstruknee}
\end{figure}

 In Figures \ref{reconstrufan} and \ref{reconstruknee}, we plot the reconstructed images of the Shepp–Logan phantom and the knee tomography, respectively.  Here, the samples of the angles  are uniformly  distributed over $[0 , \pi ]$. We consider different levels of additive order error $\eta$ as  represented in Eq.~\eqref{tomografialevelerro}.
We remark that we obtained similar results to those shown using multiple executions of our method.  To measure the effectiveness of our method, we compare the  $L^2$ error generated when our algorithm is implemented. The computed $L^2$ error is shown in  Tables \ref{taberrorfan} and \ref{taberrorknee}. 
Observing the computational error and image quality, we conclude that our reconstruction algorithm works efficiently with relatively low computational cost.

\section{Conclusions} 
In this work, we recover the gradient operator defined on Riemannian submanifolds of the Euclidean space from random samples in a neighborhood of the point of interest. Our methodology is based on the estimates of the Laplace-Beltrami operator proposed in the diffusion maps approach. The estimates do not depend on the intrinsic parametrization of the submanifold. This feature is useful in cases where it is not feasible to identify the submanifold in which the dataset is lying. \textcolor{black}{The proposed method gives a closed form of the gradient representation in the learning gradient theory. This improves the numerical implementation and the accuracy of the approximations. }A natural continuation of the present work
would be to incorporate information of the cotangent bundle and deal with a duality version of our results, in this case, the aforementioned approach would be very handy.

\begin{table}
\begin{tabularx}{1\textwidth} { 
  | >{\centering\arraybackslash}X 
  | >{\centering\arraybackslash}X 
  | >{\centering\arraybackslash}X | }
 \hline
Value of $\eta$ & With determination of the sign  & Without determination of the sign \\
 \hline
 0  & 0.0814  & 0.2087 \\
0.05  & 0.0816  &  0.2101 \\
0.1  &  0.0824  &  0.2129 \\
\hline
\end{tabularx}
\caption{Error of the reconstructed Shepp–Logan phantom. We use the $L^2$ norm to compute the errors.  Here, the sample angles are uniformly distributed over $[0 , \pi ]$.}
\label{taberrorfan}
\end{table}

\begin{table}
\begin{tabularx}{1\textwidth} { 
  | >{\centering\arraybackslash}X 
  | >{\centering\arraybackslash}X 
  | >{\centering\arraybackslash}X | }
 \hline
Value of $\eta$ & With determination of the sign  & Without determination of the sign \\
 \hline
 0  & 0.1001  & 0.1411 \\
0.05  & 0.1053  & 0.1425 \\
0.1  &  0.1114  &  0.1445 \\
\hline
\end{tabularx}
\caption{Error of the reconstructed knee tomography . We use the $L^2$ norm to compute the errors.  Here, the sample angles are uniformly distributed over $[0 , \pi ]$.}
\label{taberrorknee}
\end{table}

 Furthermore, this circle of ideas could be conjoined with the techniques proposed in Ref. \cite{PWD2020}.
\par \textcolor{black}{We conclude that the operator $\overline{P}_{t}f (x)$  locally approximates a smoothness version of the gradient of $f$. In fact, integrating by parts gives}
$$\overline{P}_{t}f (x) =  \frac{2 t^2}{d_t(x)} \left( \int_{U(x,t^\delta)} \nabla f(y) e^{\frac{-\| y-x \|^2}{2 t^2}} dy  + O( t^{\delta (d-1)}) \right ).$$
\textcolor{black}{The question of whether $\overline{P}_{t}f (x)$ is a global approximation of some smoothness gradient remains open and it could be investigated in future work.}
\par We apply our methodology in a step size algorithm as an optimization method on manifolds. This optimization method is effective in cases where it is difficult to compute the gradient of a function. As an application, we used our method to find an approximation to the sphere packing problem in dimensions 2 and 3, for the lattice packing case. Moreover, we use our approach to reconstruct tomographic images where the projected angles are unknown. The latter does not depend on {\it a  priory} knowledge of the distribution of the angles, and its execution is computationally feasible.
\par  \textcolor{black}{A natural follow-up is to apply this methodology to the dimension reduction of high-dimensional datasets}.
\par Due to the promising results obtained, another natural follow-up would be to implement our algorithm in the case of periodic lattice packing to obtain computational estimates for the sphere packing constant in several dimensions.
\par  \textcolor{black}{ In addition, we plan to implement the gradient estimates in the reinforcement learning methodology, as well as implement the proposed method for other image reconstruction problems as well as integrate with other processing techniques such as the one described in Ref.~\cite{ZMSG2003}. }

\section*{Acknowledgements}
AAG and JPZ acknowledge support from the FSU-2020-09 grant from Khalifa University. The authors acknowledge the financial support provided by CAPES,  Coordenação de Aperfeiçoamento de Pessoal de Nível Superior (Finance code 001), grant number 88887.311757/2018-00, CNPq, Conselho Nacional de Desenvolvimento Científico e Tecnológico, grant numbers 308958/2019-5 and 307873/2013-7, and FAPERJ, Fundação Carlos Chagas Filho de Amparo à Pesquisa do Estado do Rio de Janeiro, grant numbers E-26/200.899/2021 and  E-26/202.927/2017.


\bibliographystyle{siam}
\bibliography{bibli}

\newpage
\pagenumbering{arabic} 
\setcounter{page}{1} 

\begin{appendices}

\section{Numerical comparison with learning gradients}
\label{numericalcomparison}
\textcolor{black}{ In this section, we verify the consistency of Proposition \ref{estimaparamefull} and also compare the proposed algorithm  with the learning gradient approximation \cite{muksa}. We recall that given a sample set $\{x_i \}_i$  and a function $f$ in the manifold $\mathcal{M}$,  the learning gradient method computes an approximation $\vec{f}$ for the gradient using the sample points $x_i$ as}
\begin{equation}
\label{RKHS}
    \vec{f}=\sum_{i} C_i K_{t}( \cdot, x_i),
\end{equation}
\textcolor{black}{
where $K$ is the Gaussian kernel $K_{t}( x, y)=e^{-\|x-y\|^2 / 2 t^2}$, and the coefficients $C_i$ are determine by solving the optimal problem}
\begin{equation}
\label{learninggradientformula}
    \arg \max \sum_{i,j} w_{i,j} \left( f(x_j)-f(x_i)- \vec{f}(x_i) \cdot (x_j-x_i) \right)^2 +\lambda \| \vec{f} \|_{L_2}^2
\end{equation}
\textcolor{black}{
where $w_{i,j}=K(x_i,x_j)$. According to the theoretical results \cite{muksa}, to guarantee the convergence of the approximation the value for $\lambda$ is given by $\lambda=t^{d+3}$, where $d$ is the dimension of the manifold $\mathcal{M}$. The implementation of the difference between the Learning gradient and the proposed methodology lies in the fact that we compute a close form for the coefficients in the representation form \eqref{RKHS} using the Markov normalization associated with Gaussian kernels. Thus, we avoid the costly time computation  of solving the optimization problem \eqref{learninggradientformula}. }
\textcolor{black}{
We test the learning gradient and the proposed methodology to compute the gradient of the function $f: M \to \mathbb{R}$ defined as}
\begin{equation*}
    f(x)= < x , AA^{T} x>,
\end{equation*}
\textcolor{black}{
where $A$ is a squared matrix with random entries, and $<\cdot,\cdot>$ is the dot product in the Euclidean space. Here, the manifold $M$ is the curve $(c(t),c(t),c(t)) \in \mathbb{R}^9$ parameterized by}
$$ c(t)=(\cos{2 \pi t}, \sin{2 \pi t}, \cos{4 \pi t}) \in \mathbb{R}^3,$$
\textcolor{black}{
where $t \in [0,1]$   In this example, we consider random points on $t_i$ on $[0,1]$ and the set of sample points for which we compute the gradient approximation is defined as $$x_i=(c(t_i),c(t_i),c(t_i)).$$ 
We test both algorithms for different sample sizes $m$ and approximation parameters $t$. In Table \ref{tab1}, we compute the mean squared error (\textit{MSE}) of each approximation method in a logarithmic scale. We remark that since this result is probabilistic, several executions were carried out to obtain similar results without altering the conclusions concerning the tolerance of the approximation involving the several parameters.
In this experiment, we use $\delta=0.9$ and the parameter $t$ modestly small. Observe that for a fixed $t$, the \textit{MSE} error decreases when the number of sample points $m$ increases, which is consistent with the result of Proposition \ref{estimaparamefull}. In addition, observe that the proposed methodology gives a less \textit{MSE} error than the learning gradient method. This fact shows the consistency of the method with the theoretical development in this article.}

\begin{table}[H]
    \centering
    \begin{tabular}{|l|l|l|l|}
    \hline
    $t$ & $m$ & Proposed methodology & Learning gradient \\ 
    \hline
        1 & 100 & 4.13 & 4.8 \\ 
        1 & 200 & 4.04 & 5.27 \\ 
        1 & 300 & 3.8 & 5.63 \\
        1 & 400 & 3.99 & 5.07 \\ \hline
        0.5 & 100 & 3.23 & 5.51 \\ 
        0.5 & 200 & 3.69 & 5.72 \\ 
        0.5 & 300 & 3.25 & 5.4 \\ 
        0.5 & 400 & 3.41 & 5.68 \\ \hline
        0.1 & 100 & 2.45 & 5.41 \\ 
        0.1 & 200 & 2.98 & 6.38 \\ 
        0.1 & 300 & 2.66 & 5.95 \\ 
        0.1 & 400 & 2.28 & 6.02 \\ \hline
        0.05 & 100 & 3.11 & 4.93 \\ 
        0.05 & 200 & 3.28 & 5.91 \\ 
        0.05 & 300 & 2.4 & 5.1 \\ 
        0.05 & 400 & 2.14 & 5.94 \\ \hline
    \end{tabular}
    \caption{Mean squared error of the gradient approximation for the proposed method and the learning gradient in a logarithm scale. Here, $m$ is the number of sample points, $t$ is the approximation parameter}
    \label{tab1}
\end{table}

\section{Review of differential geometry}
\label{ape1}
 We review some facts of differential geometry. We refer the reader to Ref.~\cite{do1992riemannian} for a more detailed description. Given an interior point $x \in \mathcal{M}$, there exists a positive real number $\varepsilon$  such that the map $\psi=\exp_{x} \circ \,T : B(0,\varepsilon)\subset \R^d \to \mathcal{M}$ is a local chart. Here, $\exp_{x}$ is the exponential map at the point $x$, and $T:\R^d \to T_{x} \mathcal{M}$ is a rotation from $\R^d$ onto $T_{x} \mathcal{M}$, both sets considered subsets of $\R^n$. The chart $\psi$ defines the normal coordinates at point $x$. 
 \par Given a smooth function $f \in C^{\infty} (\mathcal{M})$, the gradient operator  $\nabla f (x)\in T_{x} \mathcal{M}$  is given in  normal coordinates by
$$ \nabla f(x)= \sum_{i=1}^d \frac{\partial f}{\partial x_i} T(e_i).$$
Here, $e_i$ is the standard basis in $\R^d$. Now, we recall some estimates that use normal coordinates that are useful when estimating approximations for differential operators. The Taylor series of $\psi$ around the point $0$ is given by
\begin{equation}
    \psi(v)= x+ T (v)+\frac{1}{2} D^2\psi_0(v,v)+ O(\|  v\|^3).
    \label{taylorexpo}
\end{equation}
Let $v \in B(0,\varepsilon)\subset \R^d$, and consider the geodesic $\gamma_{T(v)}$,  with initial tangent vector $T(v) \in T_{x} \mathcal{M}$, then using Estimate~(\ref{taylorexpo}) we obtain
$$\gamma_{T(v)}(t)=x+T(v)\,t+\frac{1}{2} D^2\psi_0\,(v,v) t^2+O(\|v\|^3)t^3. $$
Since the covariant derivative of a geodesic vanishes, then $\gamma''_{{T(v)}}$ is orthogonal to $T_{x} \mathcal{M}$. Thus, we have the following estimates
\begin{equation}
\|\psi(v)-x\|^2= \|T(v)\|^2+O(\|v\|^4),
\label{estimativaorden2}
\end{equation}
and
\begin{equation}
   \mathcal{P}_x(\psi(v)-x)= T(v)+O(\|v\|^3),
  \label{estimativaorden3}
\end{equation}
where $\mathcal{P}_x$ is the orthogonal projection on  $T_{x} \mathcal{M}$.
Using the Estimates~(\ref{estimativaorden2}) and~(\ref{estimativaorden3}), we obtain that there exist positive constants $M_1$ and $M_2$ such that for $\|v\|$ small
$$\|v\|-M_2 \|v\|^3 \le \|\psi(v)-x\|\le M_1 \|v\|.$$
Thus, if $\|v\|^2 \le \frac{1}{2M2}$ we have
$$\frac{1}{2}\|v\|  \le \|\psi(v)-x\|\le M_1 \|v\|.$$
This says that for $t$ small
\begin{equation}
    \label{condilocal}
    B(0,t/{M_1}) \subseteq \psi^{-1} (U(x,t^\delta)) \subseteq B(0,2 t).
\end{equation}

\section{Expansion of the gradient operator}
\label{ape2}
Here, we show the technical details of the proof of Theorem \ref{teo1}. The main idea is to use the Taylor expansion of the function $f$ around the point $x$.
\begin{lem}
\label{lemaprinci}
Assume that $\frac{1}{2} < \delta<1$, and let $K:\mathcal{M} \times \mathcal{M} \to \R^m$  be a vector value kernel. Define
$$P_{t}(x)=\int_{U(x,t^\delta)} K(x,y) \, e^{\frac{-\| y-x \|^2}{2 t^2}}   dy,$$
where $U(x,t^\delta)$ is defined as in Eq.~(\ref{conjuntopequ}). Assume that for $t$ small, the function $\psi:B(0,2t^\delta) \to \mathcal{M} $ defines normal coordinates in a neighborhood of $x$, and let $S$ be a vector value function defined in $\R^d$ such that
$$K(x,\psi(v))-S(v)=O(\|v\|^r),$$
and
$$K(x,y)=O(\|x-y\|^s).$$
Then, we have
$$ P_{t}(x)=O(   (e^{C_2 t^{4\delta-2}} -1) t^{s+d}+ t^{r+d})+\int_{\psi^{-1} (U(x,t^\delta))} S(v) \, e^{\frac{-\| T(v)\|^2}{2 t^2}} dv.$$
\end{lem}
\begin{proof}
Using Eq.~(\ref{condilocal}), we assume that for $t$ small, the set $U(x,t^\delta)$ lies in the image of a normal chart $\psi:B(0,2t^{\delta}) \to \mathcal{M} $ centered in $x$. Thus,
$$\begin{array}{rcl}  \int_{U(x,t^\delta)} K(x,y) \,  e^{\frac{-\| y-x \|^2}{2 t^2}} dy & = & \int_{\psi^{-1} (U(x,t^\delta))} K(x,\psi(v)) e^{\frac{-\| \psi(v)-x \|^2}{2 t^2}} dv \\ \, & = & \int_{\psi^{-1} (U(x,t^\delta))}  K(x,\psi(v)) (e^{\frac{-\| \psi(v)-x \|^2}{2 t^2}}-e^{\frac{-\| T(v) \|^2}{2 t^2}}) dv \\ \, & + & \int_{\psi^{-1} (U(x,t^\delta))} (K(x,\psi(v))-S(v)) e^{\frac{-\| T(v)\|^2}{2 t^2}} dv
\\  \, & + & \int_{\psi^{-1} (U(x,t^\delta))} S(v) \, e^{\frac{-\| \T(v)\|^2}{2 t^2}} dv.
\end{array}  $$
We now estimate
$$A=\int_{\psi^{-1} (U(x,t^\delta))} K(x,\psi(v)) (e^{\frac{-\| \psi(v)-x \|^2}{2 t^2}}-e^{\frac{-\| T(v) \|^2}{2 t^2}}) dv. $$
Using Eq.~(\ref{estimativaorden2}), and the inequality $ |e^x-1|\le e^{|x|}-1$ we obtain
$$\begin{array}{rcl}  |e^{\frac{-\| \psi(v)-x \|^2}{2 t^2}}-e^{\frac{-\| T(v) \|^2}{2 t^2}}| & = & e^{\frac{-\| T(v) \|^2}{2 t^2}} | e^{\frac{O(\|v\|^4)}{2 t^2}} -1| \\ \, & \le & e^{\frac{-\| T(v) \|^2}{2 t^2}} ( e^{\frac{C_1\|v\|^4}{2 t^2}} -1).
\end{array}  $$
Therefore, by Equation~(\ref{condilocal}) we obtain 
$$\begin{array}{rcl} \| A\| & \le & C_3 \, \, t^s ( e^{C_2 t^{4\delta-2}} -1) t^d \int_{\R^d} \| v \|^s e^{-\| v \|^2 /2}  dv \\ \, & = & O( (e^{C_2 t^{4\delta-2}} -1) t^{s+d} ).
\end{array}  $$
On the other hand, by assumption we have 
$$\int_{\psi^{-1} (U(x,t^\delta))} ( K(x,\psi(v))-S(v)) e^{\frac{-\| T(v)\|^2}{2 t^2}} dv=O( t^{r+d} \,).$$
\end{proof}
\begin{lem} \label{lemma2}
Under the same assumptions of  Lemma \ref{lemaprinci}, we define 
$$E=\int_{\psi^{-1} (U(x,t^\delta))} Q(v) e^{\frac{-\| T(v)\|^2}{2 t^2}} g(v)dv,$$
where $g$ is a smooth function and  $Q$ is a homogeneous polynomial of degree $l$. Then, we have
$$E=\int_{\R^d} Q(v) e^{\frac{-\| T(v)\|^2}{2 t^2}} (g(0)+\sum \frac{\partial g}{\partial v_i} (0) v_i) dv\,+ O(t^{d+l} e^{-M_2 t^{2(\delta-1)}}+t^{d+2+l}).$$
\end{lem}

\begin{proof}
Using the Taylor expansion of $g$ around $0$ we have
$$E=\int_{\psi^{-1} (U(x,t^\delta))} Q(v) e^{\frac{-\| T(v)\|^2}{2 t^2}} (g(0)+\sum \frac{\partial g}{\partial v_i} (0)\,   v_i+ O(\|v\|^2) )dv .$$
Let $B$ be defined as
$$ B= \|\int_{\R^d \setminus \psi^{-1} (U(x,t^\delta))} Q(v) e^{\frac{-\| T(v)\|^2}{2 t^2}} (g(0)+\sum \frac{\partial g}{\partial v_i} (0)\,  v_i) dv\|.$$
Using Eq.~(\ref{condilocal}) and the fast decay of the exponential function, we obtain that
$$ B \le  C_4  t^{d+l} e^{-M_2 t^{2(\delta-1)}} \int_{\R^d \setminus B(0 ,  t^{\delta-1}/M_1)}  P(\|v\|) e^{\frac{-\| T(v)\|^2}{4}} dv.$$
for a certain polynomial $P$. Therefore, we have
$$B = O(t^{d+l} e^{-M_2 t^{2(\delta-1)}} ),$$
for a proper constant $M_2$. Finally, we observe that
$$\int_{\psi^{-1} (U(x,t^\delta))} Q(v) e^{\frac{-\| T(v)\|^2}{2 t^2}} O(\|v\|^2) dv=O(t^{d+2+l}  ).$$
\end{proof}
We recall the following computations related to the moments of the normal distribution that are useful in proving  Theorem~\ref{teo1}.
For all index $i$ 
$$\int_{\R^d} v_i e^{\frac{-\| T(v)\|^2}{2 t^2}}dv=0,$$
and
$$\int_{\R^d} v_i^2 e^{\frac{-\| T(v)\|^2}{2 t^2}}dv=(2 \pi)^{\frac{d}{2}}\, t^{d+2},$$
moreover, if $i \neq j$ then
$$\int_{\R^d} v_i \, v_j e^{\frac{-\| T(v)\|^2}{2 t^2}}dv=0.$$

\begin{lem} 
\label{lemanor} 
Under the same assumptions of Lemmas \ref{lemaprinci} and \ref{lemma2} we have
\begin{equation}
    \label{estimativanormali}
    d_t(x)= (2 \pi)^{\frac{d}{2}} t^d+O(t^{d+4 \delta -2}).
\end{equation}

\end{lem}
\begin{proof} 
We apply Lemmas \ref{lemaprinci} and \ref{lemma2} to the functions $K(x,y)=1$, $S(v)=1$, $Q(v)=1$, and $g(x)=1$. We use the parameters $r=2$, $s=0$ and $l=0$. Using the exponential decay we obtain the following estimate
$$d_t(x)= (2 \pi)^{\frac{d}{2}} t^d+O(t^{d+4 \delta -2}).$$
\end{proof}
\begin{proof}[Proof of Theorem \ref{teo1}]
We apply Lemmas \ref{lemaprinci} and \ref{lemma2} to the functions $K(x,y)= (y-x)(f(y)-f(x))$, $S(v)=T(v)(f(\psi(v))-f(x))=\sum v_i (f(\psi(v))-f(x)) T(e_i) $ , $Q(v)=v_i $ and $g(v)=(f(\psi(v))-f(x))$. Since $\psi(v)-x-T(v)=O(\|v\|^2)$ and $f(\psi(v))-f(x)=O(\|v\|^1)$,  then the parameters that we use are $r=3$, $s=2$ and $l=1$. Again, using the exponential decay we have that
\begin{equation}
    \int_{U(x,t^\delta)} \overline{K}(x,y) \, e^{\frac{-\| y-x \|^2}{2 t^2}} dy=(2 \pi)^{\frac{d}{2}}\, t^{d+2}\sum \frac{\partial f}{\partial v_i} (0)\,   T(e_i)+O(t^{d+4 \delta}).
    \label{ecuafinal}
\end{equation}
Finally we use Eq.~(\ref{estimativanormali}) of Lemma \ref{lemanor} to conclude the result.
\end{proof}
\end{appendices}

\medskip
Received xxxx 20xx; revised xxxx 20xx.
\medskip

\textit{E-mail address}, Alvaro Almeida Gomez: \texttt{alvaro.gomez@ku.ac.ae}
\medskip

\textit{E-mail address}, Antônio J.  Silva Neto: \texttt{ajsneto@iprj.uerj.br}
\medskip

\textit{E-mail address}, Jorge P. Zubelli: \texttt{zubelli@gmail.com}
\medskip

\end{document}